\def\LoRAForgeArxiv{1}
\documentclass{article}

\usepackage{latexsym}

\usepackage[utf8]{inputenc}
\usepackage[T1]{fontenc}
\PassOptionsToPackage{hyphens}{url}
\usepackage{hyperref}
\ifdefined\LoRAForgeArxiv
  \usepackage[accepted]{mlsys2025}
  \makeatletter
  \renewcommand{\Notice@String}{Preprint.}
  \makeatother
\else
  \usepackage{mlsys2025}
\fi
\usepackage{url}
\usepackage{booktabs}
\usepackage{amsmath,amssymb,mathtools}
\usepackage{amsthm}
\usepackage{bm}
\newcommand{\mat}[1]{\bm{#1}}

\usepackage{microtype}
\usepackage{xcolor}
\usepackage{graphicx}
\usepackage{tikz}
\usetikzlibrary{arrows.meta,calc,shadows,shapes.geometric,decorations.pathreplacing}
\usepackage{caption}
\usepackage{subcaption}
\usepackage{tabularx}
\usepackage{multirow}
\usepackage{colortbl}
\usepackage{pifont}
\usepackage{placeins}
\usepackage{enumitem}
\newcommand{\lfuniform}{LoRAForge-U}
\newcommand{\lfspectral}{LoRAForge-S}

\definecolor{navy}{HTML}{17365D}
\definecolor{linkblue}{HTML}{3F79B7}
\definecolor{oraclebg}{HTML}{F1F5F9}
\definecolor{bestbg}{HTML}{D4EDDA}
\definecolor{resultours}{RGB}{232,240,252}
\definecolor{resultgroup}{gray}{0.91}
\definecolor{resulthead}{gray}{0.96}
\hypersetup{colorlinks=true,citecolor=navy,linkcolor=navy,urlcolor=linkblue}

\mlsystitlerunning{LoRAForge: Co-Designing Post-Hoc Compression and Execution}
\hypersetup{pdftitle={LoRAForge: Co-Designing Post-Hoc Compression and Execution for Efficient Multi-LoRA Serving},pdfsubject={MLSys 2027 Submission}}

\ifdefined\LoRAForgeArxiv
  \usepackage{fontawesome5}
  \newcommand{\LoRAForgePDFAuthors}{Shenghui Li and Thiemo Voigt}

  \hypersetup{pdfsubject={Preprint},pdfauthor={\LoRAForgePDFAuthors}}
\fi

\begin{document}

\twocolumn[
\mlsystitle{LoRAForge: Co-Designing Post-Hoc Compression and Execution for Efficient Multi-LoRA Serving}
\ifdefined\LoRAForgeArxiv
  \begin{mlsysauthorlist}
\mlsysauthor{Shenghui Li}{uppsala}
\mlsysauthor{Thiemo Voigt}{uppsala,rise}
\par\vspace{0.35em}
{\normalsize\hypersetup{urlcolor=navy}\href{https://github.com/lishenghui/loraforge}{\faGithub\hspace{0.4em}\nolinkurl{github.com/lishenghui/loraforge}}}
\end{mlsysauthorlist}
\mlsysaffiliation{uppsala}{Uppsala University, Uppsala, Sweden}
\mlsysaffiliation{rise}{Research Institutes of Sweden, Stockholm, Sweden}
\mlsyscorrespondingauthor{Shenghui Li}{shenghui.li@angstrom.uu.se}
\mlsyscorrespondingauthor{Thiemo Voigt}{thiemo.voigt@angstrom.uu.se}

\else
\begin{mlsysauthorlist}
\mlsysauthor{Anonymous Authors}{anonymous}
\end{mlsysauthorlist}
\mlsysaffiliation{anonymous}{Anonymous Institution}
\mlsyscorrespondingauthor{}{}
\fi
\mlsyskeywords{LoRA, compression-execution co-design, multi-adapter serving, rank allocation}
\vskip 0.3in

\begin{abstract}
Multi-LoRA serving allows many task-specific adapters to share a common base model, but scaling to large adapter fleets introduces substantial storage, loading, and execution overheads. Post-hoc compression can reduce the memory footprint of trained LoRA adapters, yet smaller ranks do not automatically translate into faster serving. For instance, padding, factor access, and generic kernels can obscure the retained low-rank structure, while independently recompressing thousands of modules also incurs significant preparation overhead. We present LoRAForge, a system connecting batched factor
construction to rank-specialized execution through an explicit retained-rank
representation. It supports two allocation policies: uniform retention
ratios (\lfuniform{}) and spectral-energy budget allocation (\lfspectral{}).
Both share FlashTSQR factor construction and the same serving backend.
FlashTSQR exploits parallelism across modules and row tiles, accelerating
factor construction by up to $2.28\times$ over matched library QR.
In a fixed-work Qwen control, the shared runtime adds $1.17$--$1.22\times$ throughput
across both compressed policies, with negligible gains on original adapters.
Across 85  trained open-weighted Qwen and Llama adapters, the spectral policy's floored quality
configuration retains about 7.7\% of adapter parameters with fleet-average
score drops of at most 1.2 points on a 0--100 scale. The same rank-floored $1/12$ spectral exports achieve substantial token-throughput gains on both AMD and NVIDIA GPUs.
\end{abstract}
]
\printAffiliationsAndNotice{}

\section{Introduction}
\label{sec:intro}

Low-Rank Adaptation (LoRA)~\citep{hu2022lora} is a widely 
adopted parameter-efficient fine-tuning method that adapts large language models (LLMs) to specific tasks by training only a small set of adapter parameters while keeping the base model frozen. Recently, multi-LoRA serving has emerged as an increasingly popular paradigm for consolidating many such task-specific adapters on shared model infrastructure~\citep{nguyen2024multitenantlora,anyscale2026multilora}. Rather than provisioning a separate base model replica for each specialization, the service shares the base weights and activates the adapter specified by each inference request.
This paradigm supports a wide range of applications, including question answering~\citep{lazier2025aclora,chen2025svrag}, multilingual machine translation~\citep{nguyen2024multitenantlora,dong2025mlaslora}, and multi-agent systems~\citep{jeon2026lragent}.

Although individual LoRA adapters are compact, their aggregate memory footprint can become substantial when dozens to hundreds of adapters share a GPU,
consuming GPU memory and competing with the KV cache for capacity.
Existing multi-LoRA serving systems commonly mitigate this pressure by keeping
only a subset of adapters resident on the GPU and loading nonresident adapters
from host memory on demand~\citep{chen2024punica,sheng2024slora,li2025toppings}.
When the active working set exceeds GPU adapter capacity, requests for
nonresident adapters can trigger repeated eviction and host-to-device loading.
These systems seek to hide such transfers by overlapping asynchronous loading
or prefetching with computation~\citep{chen2024punica,sheng2024slora}.
However, when transfer times exceed the available computation overlap or
aggregate transfer demand saturates the host--device link, loading delays
enter the execution critical path, increasing request latency and reducing
serving throughput~\citep{li2025toppings}. %

A growing line of research exploits redundancy in LoRA ranks at different stages and granularities. Training-time methods such as AdaLoRA and DyLoRA adapt rank allocation during fine-tuning~\citep{zhang2023adalora,valipour2023dylora}, making them unsuitable for arbitrary, already-trained adapter fleets. Post-hoc methods such as PARA compress individual trained adapters without retraining~\citep{kumaravelu2026para}, but do not allocate a shared compression budget across adapters. At the fleet level, Compress then Serve jointly compresses adapters using
shared bases and adapter-specific scaling matrices~\citep{brueelgabrielsson2024compressthenserve}.
However, its compression procedure requires iterative joint fitting,
repeatedly processing the adapter collection. It also uses a common
compressed rank within each shared-basis group, rather than allocating
ranks individually across adapters under a fleet-wide budget.

Turning post-hoc compression into an efficient serving solution presents two further challenges. {First, compression itself must scale to an entire adapter fleet.} Spectral methods provide the mathematical tools for rank reduction~\citep{kumaravelu2026para,li2026fraq}, but naively applying library QR/SVD routines independently to thousands of LoRA factors incurs substantial intermediate storage and kernel-launch overhead. Efficient fleet-scale compression therefore requires batching these operations across adapters and modules. 
{Second, lower ranks do not automatically translate into proportional serving gains.} In the default LoRA implementations we examine in vLLM~\citep{kwon2023vllm} and SGLang~\citep{zheng2024sglang}, fixed or coarsely tiled rank dimensions can waste work at the small ranks produced by compression (Appendix~\ref{app:execution-implementation}). Realizing speedups therefore requires efficiently executing the resulting smaller matrix operations during both prompt processing and token generation.

We propose \textbf{LoRAForge}, a system that makes compressed LoRA adapters
efficient to construct and serve. Its retained-rank representation connects
ordered factor prefixes to loading, factor access, and kernel dispatch,
allowing a shared runtime to exploit ranks chosen by different policies.
\lfuniform{} retains a uniform fraction of each module's original rank;
\lfspectral{} allocates a shared parameter budget by spectral energy.
Both use our compression primitive FlashTSQR for batched factor construction and the same shrink and
expand backend. Our matched-work controls show execution gains for both
policies, while the spectral policy offers a measured quality--resource
tradeoff. With spectral allocation, LoRAForge achieves $1.59\times$ and
$1.92\times$ the token throughput of uncompressed native serving on MI250X
and GH200, respectively. Our main contributions are:
\begin{itemize}[leftmargin=1.5em,labelsep=0.5em]
\item
We propose a budget-aware fleet compression pipeline supporting uniform
retention and spectral allocation. We develop an efficient GPU primitive,
FlashTSQR, that accelerates batched factor
construction, while reusable ordered factors and retained-rank metadata
provide a common serving representation.
\item
We design rank-specialized execution that preserves live directions through
padding and loading. Compact factor access, specialized shrink/expand kernels,
and phase-aware dispatch translate the retained ranks into efficient serving.
\item
We evaluate LoRAForge across 85 trained open-weight adapters on AMD and NVIDIA
GPUs, measuring compression cost, task quality, and serving performance.
Paired Qwen controls separate shared execution gains from allocation and
resource tradeoffs.
\end{itemize}

\section{Background and Motivation}
\label{sec:background}

\begin{figure*}[t]
    \centering
    \begin{subfigure}[t]{0.49\textwidth}
        \centering
        \includegraphics[width=\linewidth,clip]{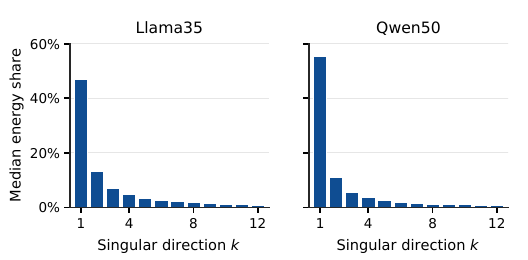}
        \caption{Energy distribution}
    \end{subfigure}\hfill
    \begin{subfigure}[t]{0.49\textwidth}
        \centering
        \includegraphics[width=\linewidth,clip]{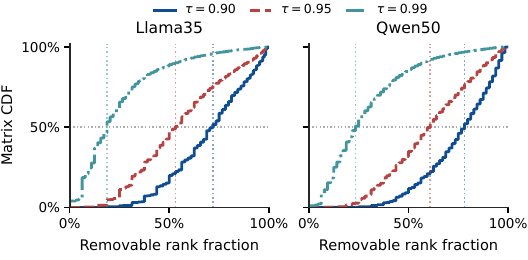}
        \caption{Removable rank}
    \end{subfigure}
    \captionsetup{font=footnotesize}
    \caption{\textbf{Rank redundancy in our two task-adapter fleets.} Llama-3.1-8B-Instruct: 35 adapters and 7,840 LoRA matrices; Qwen3-4B-Instruct-2507: 50 adapters and 12,492 matrices. Energy panels (a) show the median normalized squared singular values; rank panels (b) show the CDF of removable-rank fractions at three energy thresholds, with dotted vertical lines marking the medians. Matrices are weighted equally within each fleet, and corresponding panels use shared axis ranges. At $\tau=.90$, median retained-rank fraction is $28.1\%$ for Llama and $21.9\%$ for Qwen.}
    \label{fig:rank-utilization}
\end{figure*}

\paragraph{Post-Hoc LoRA Compression}
For a pretrained weight matrix
$\mat W_{0,\ell}\in\mathbb{R}^{d_{\mathrm{out}}\times d_{\mathrm{in}}}$,
LoRA keeps $\mat W_{0,\ell}$ frozen and learns an additive low-rank update
$
  \Delta\mat W_{i,\ell}
  =
  \frac{\alpha_{i,\ell}}{r_{i,\ell}}
  \mat B_{i,\ell}\mat A_{i,\ell},
  \label{eq:lora-update}
$
where
$\mat B_{i,\ell}\in\mathbb{R}^{d_{\mathrm{out}}\times r_{i,\ell}}$,
$\mat A_{i,\ell}\in\mathbb{R}^{r_{i,\ell}\times d_{\mathrm{in}}}$,
and $r_{i,\ell}$ is the nominal rank \citep{hu2022lora}.
An adapter consists of such updates across a collection of modules while
sharing the same frozen base model. An already-trained LoRA module can be further compressed without retraining by
reducing the rank of its effective update $\Delta\mat W_{i,\ell}$. Let its
compact singular value decomposition be
$
  \Delta\mat W_{i,\ell}
  =
  \mat U_{i,\ell}
  \mat\Sigma_{i,\ell}
  \mat V_{i,\ell}^{\top},
  \label{eq:lora-svd}
$
with singular values
$\sigma_{i,\ell,1}\ge\cdots\ge\sigma_{i,\ell,r_{i,\ell}}$.
Retaining the leading $k$ singular directions gives
$
  \Delta\mat W_{i,\ell,k}
  =
  \mat U_{i,\ell,:k}
  \mat\Sigma_{i,\ell,:k}
  \mat V_{i,\ell,:k}^{\top},
  \label{eq:lora-truncation}
$
the optimal rank-$k$ approximation under the Frobenius norm. The truncated
update can be refactored into rank-$k$ LoRA factors and served in the same form
as the original module. Such SVD recompression is used in post-hoc adapter
compression \citep{brueelgabrielsson2024compressthenserve,kumaravelu2026para}.
It can also be computed directly from low-rank factors or compact coordinate
matrices without materializing the full update, substantially reducing the
recompression cost \citep{ramesh2026florist,ramesh2026spectral,li2026fraq}. A common spectral compression rule selects the smallest retained rank whose
leading singular directions preserve a prescribed fraction $\tau$ of the
spectral energy,
$
  k_{i,\ell}(\tau)
  =
  \min\left\{
  k\in\{1,\ldots,r_{i,\ell}\}:
  \frac{\sum_{j=1}^{k}\sigma_{i,\ell,j}^{2}}
       {\sum_{j=1}^{r_{i,\ell}}\sigma_{i,\ell,j}^{2}}
  \ge \tau
  \right\}.
  \label{eq:spectral-rank}
$
The evaluated allocator retains at least one direction per targeted module;
it does not remove LoRA branches entirely.

\paragraph{Multi-LoRA Serving.}
During inference, a LoRA update is applied as two low-rank operations.
For the token rows $\mat X$ assigned to one adapter, the shrink projection computes
$\mat Z=\mat X\mat A^\top$, reducing the feature dimension to the adapter rank.
The expand projection updates the base-model output as
$\mat Y\mathrel{+}=s\mat Z\mat B^\top$, where $s$ is the LoRA scaling factor.
Requests in the same batch may select different adapters, so serving systems
group or segment token rows by adapter identity and execute the corresponding
LoRA factors alongside the shared base-model computation~\citep{chen2024punica,sheng2024slora}.
Adapter factors remain resident on the GPU or are loaded on demand when
the active working set exceeds the available adapter memory~\citep{sheng2024slora,li2025toppings}.
Consequently, serving efficiency depends on both the stored parameters and
the ranks and shapes of the low-rank operations exposed to the execution backend.

\paragraph{Motivation.}
\label{sec:motivation}

Our spectral characterization uses the same 35 Llama-3.1-8B-Instruct and
50 Qwen3-4B-Instruct-2507 adapters as quality evaluation and serving,
totaling 20,332 LoRA matrices.
At 90\% retained squared singular energy, the median matrix needs only
28.1\% and 21.9\% of its nominal rank, respectively
(Fig.~\ref{fig:rank-utilization}). These observations motivate post-hoc
compression, but energy concentration alone does not guarantee task quality.
Uniform retention and spectral reallocation turn this redundancy into
smaller ranks with different quality and resource costs. Both require
execution to exploit the retained structure. In our fixed-output Qwen
control, the same optimized path adds $1.18$--$1.22\times$ throughput for
uniform retention and $1.17$--$1.19\times$ for spectral allocation, but
only $0.99$--$1.00\times$ on original weights
(Figure~\ref{fig:allocation-execution}). These results motivate a shared
compression--execution interface rather than tying serving gains to one
allocator. Effective ranks must remain explicit through loader padding,
factor access, and dispatch; the runtime then selects execution paths for
the resulting rank, shape, and serving phase.

\section{LoRAForge Overview}
\label{sec:overview}

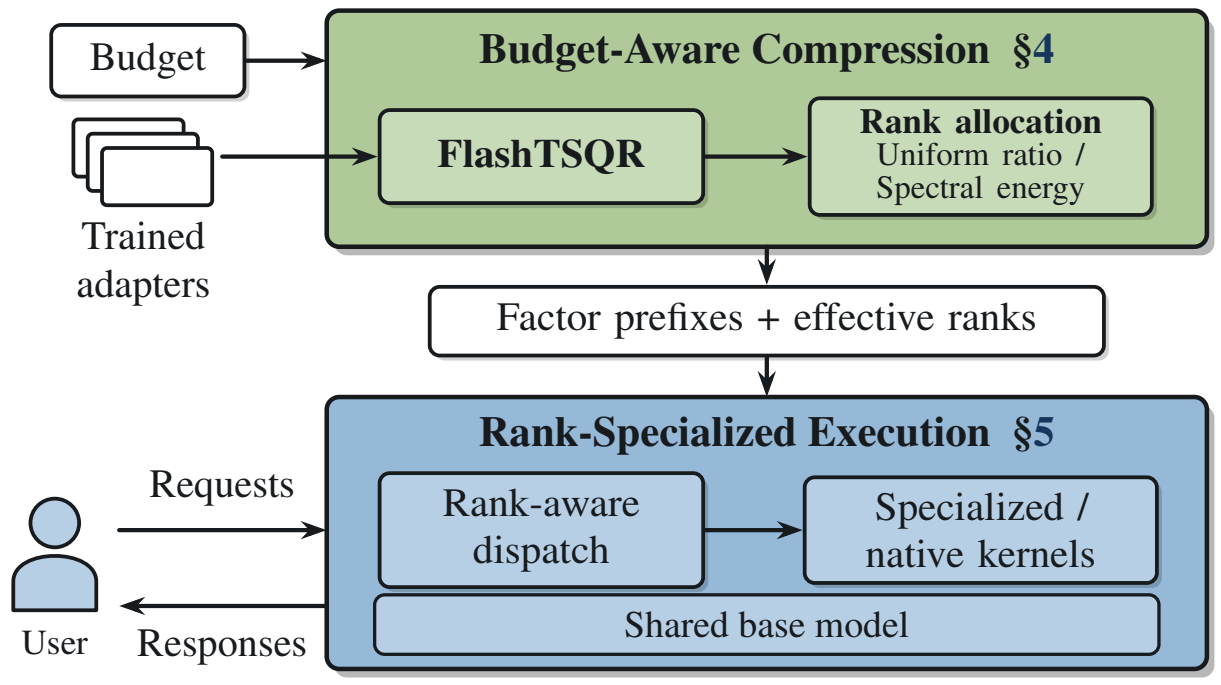
\begin{figure}[t]
\centering
\resizebox{\linewidth}{!}{\begingroup
\definecolor{ovblue}{HTML}{7CA8CE}
\definecolor{ovgreen}{HTML}{9BBD7E}
\definecolor{ovink}{HTML}{171B1E}
\begin{tikzpicture}[
 x=1cm,y=1cm,text=ovink,font=\fontsize{8}{9.5}\selectfont,
 >={Stealth[length=1.8mm,width=1.25mm]},
 flow/.style={->,draw=ovink,line width=.85pt,rounded corners=2pt,line cap=round,line join=round},
 io/.style={flow},
 panel/.style={draw=ovink,line width=.9pt,rounded corners=2.5pt,
  drop shadow={shadow xshift=1.5pt,shadow yshift=-1.7pt,fill=black,opacity=.28}},
 component/.style={draw=ovink,line width=.75pt,rounded corners=2pt,
  drop shadow={shadow xshift=1pt,shadow yshift=-1pt,fill=black,opacity=.18},
  align=center,minimum height=.65cm,inner sep=3pt},
 head/.style={font=\fontsize{8.5}{10}\selectfont\bfseries,align=center},
 label/.style={align=center,inner sep=1pt}
]
\node[component,fill=white,minimum width=1.35cm,minimum height=.5cm]
 (budget) at (.9,4.5) {Budget};
\draw[draw=ovink,fill=white,line width=.75pt,rounded corners=1pt]
 (.38,3.69) rectangle (1.15,4.09);
\draw[draw=ovink,fill=white,line width=.75pt,rounded corners=1pt]
 (.48,3.59) rectangle (1.25,3.99);
\draw[draw=ovink,fill=white,line width=.75pt,rounded corners=1pt]
 (.58,3.49) rectangle (1.35,3.89);
\node[label] at (.87,3.08) {Trained\\adapters};

\draw[panel,fill=ovgreen!80] (2.15,3.2) rectangle (8.3,4.85);
\node[head] at (5.225,4.56)
 {Budget-Aware Compression\enspace\S\ref{sec:method}};
\node[component,fill=ovgreen!55,text width=2.05cm]
 (flash) at (3.65,3.83) {\textbf{FlashTSQR}};
\node[component,fill=ovgreen!55,text width=2.25cm,inner sep=2pt]
 (allocation) at (6.72,3.83) {\shortstack{
 {\fontsize{7}{8}\selectfont\bfseries Rank allocation}\\[-1pt]
 {\fontsize{6.5}{7.5}\selectfont Uniform ratio /}\\[-1pt]
 {\fontsize{6.5}{7.5}\selectfont Spectral energy}}};
\draw[flow] (budget.east) -- (2.15,4.5);
\draw[flow] (1.42,3.83) -- (flash.west);
\draw[flow] (flash.east) -- (allocation.west);

\node[component,fill=white,minimum height=.48cm,text width=4.5cm]
 (compressed) at (5.225,2.68) {Factor prefixes + effective ranks};
\draw[flow] (5.225,3.2) -- (compressed.north);
\draw[flow] (compressed.south) -- (5.225,2.15);

\draw[panel,fill=ovblue!80] (2.15,.25) rectangle (8.3,2.15);
\node[head] at (5.225,1.88)
 {Rank-Specialized Execution\enspace\S\ref{sec:serving}};
\node[component,fill=ovblue!55,text width=2.05cm]
 (dispatch) at (3.65,1.22) {Rank-aware\\dispatch};
\node[component,fill=ovblue!55,text width=2.25cm]
 (kernels) at (6.72,1.22) {Specialized /\\native kernels};
\draw[flow] (dispatch.east) -- (kernels.west);
\node[draw=ovink,fill=ovblue!55,line width=.65pt,
 rounded corners=2pt,minimum width=5.48cm,minimum height=.38cm,
 font=\fontsize{7.5}{9}\selectfont]
 at (5.225,.56) {Shared base model};

\draw[draw=ovink,fill=ovblue!55,line width=.75pt]
 (.25,1.27) circle[radius=.17];
\draw[draw=ovink,fill=ovblue!55,line width=.75pt,line join=round]
 (-.04,.66) -- (-.04,.86)
 .. controls (-.04,1.1) and (.54,1.1) .. (.54,.86)
 -- (.54,.66) -- cycle;
\node[label,font=\fontsize{7}{8.2}\selectfont] at (.25,.43) {User};
\node[label] at (1.42,1.51) {Requests};
\draw[io] (.7,1.22) -- (2.15,1.22);
\draw[io] (2.15,.69) -- (.7,.69);
\node[label] at (1.42,.4) {Responses};
\end{tikzpicture}
\endgroup}
\caption{\textbf{LoRAForge Overview.}
Two rank-allocation policies share factor construction and a retained-rank
interface to the same serving runtime.}
\label{fig:lorafleet-overview}
\end{figure}

LoRAForge takes independently trained adapters for a shared base model and
serves requests selecting adapters by identity. Figure~\ref{fig:lorafleet-overview}
connects preparation and serving through ordered factor prefixes and explicit
module ranks. The system supports \lfuniform{} and \lfspectral{} as two
allocation policies, while preserving each adapter's independent factorization.

During preparation, FlashTSQR constructs reusable ordered factors and spectra.
Uniform retention selects a common fraction of each module's original rank;
spectral allocation distributes a compact-parameter budget by retained energy
and direction cost. Both export factor prefixes and effective ranks, even
when checkpoint storage requires padding. During serving, the shared runtime
propagates these ranks into factor access and dispatch. It selects specialized
or native shrink/expand kernels by rank, shape, and phase, applying the chosen
adapter alongside the base model. Dispatch changes how factors execute,
not which directions they retain. Sections~\ref{sec:method} and~\ref{sec:serving}
describe this shared preparation and execution design.

\section{Budget-Aware Fleet Compression}
\label{sec:method}

LoRAForge separates rank selection from factor construction and execution.
We first define their retained-rank contract, then the two allocation policies
and their shared FlashTSQR construction pipeline.

\subsection{The Retained-Rank Contract}
\label{sec:retained-rank}

Both allocation policies produce the same retained-rank representation:
ordered factor prefixes together with each module's effective rank.
The contract requires the serving system to preserve the distinction between
{effective rank} and {physical capacity}: padding for storage or
kernel compatibility must not redefine which directions are live.

For each module, the manifest associates its adapter/module identity with
retained rank $k_m$, selected factor prefixes, and update scale. The loader
propagates $k_m$ to device metadata for factor access and kernel dispatch,
even when checkpoint tensors are padded to an adapter-wide maximum.
Exports preserve the effective update; if scaling is absorbed into the
factors, loader scaling is set to one.

The runtime may repack factors, fuse projections, or bucket ranks, but must
preserve the exported update and distinguish live directions from padding.
Ranks are fixed before evaluation: dispatch determines {how to execute}
the retained factors, not {which directions to retain}.
Section~\ref{sec:serving} implements this contract through effective-rank
bounds, compact factor access, and phase-, rank-, and shape-aware dispatch.

\subsection{Two Rank-Allocation Policies}
\label{sec:allocation}

For module $m$, let $r_m$ be its original rank,
$\sigma_{m,j}$ its descending effective singular values (including LoRA scale),
and $c_m=d_{\mathrm{in},m}+d_{\mathrm{out},m}$ the parameter cost per direction.
Both policies retain leading singular directions; they differ in how many.
\lfuniform{} uses $u_m=\max(1,\lfloor r_m/d\rfloor)$ for a common divisor $d$.
This is a uniform retention ratio, not necessarily equal final ranks.
Its cost $P_U=\sum_m c_m u_m$ defines the matched compact budget.
\lfspectral{} reallocates this budget according to spectral energy:
\begin{equation}
\begin{aligned}
 \max_{\{k_m\}}\quad &\sum_m\sum_{j=1}^{k_m}\sigma_{m,j}^{2},\\[3pt]
 \text{s.t.}\quad &\sum_m c_m k_m\le P,\\[2pt]
 &\ell_m\le k_m\le r_m,\quad k_m\in\mathbb Z.
\end{aligned}
\label{eq:spectral-fleet}
\end{equation}
Here $P=P_U$ and $\ell_m$ is the configured rank floor. The objective
minimizes summed squared Frobenius residuals, so larger-norm adapters can
receive more budget. It is a reconstruction proxy, not a latency or quality
guarantee; Section~\ref{sec:quality-budgets} evaluates the policy tradeoff.

\paragraph{Cost-aware prefix selection.}
We initialize $k_m=\ell_m$, requiring $P$ to cover these floors, and order
remaining directions by marginal energy per parameter,
\begin{equation}
 \eta_{m,j}=\frac{\sigma_{m,j}^2}{c_m},\qquad j>\ell_m.
 \label{eq:energy-density}
\end{equation}
A direction is admitted if its energy is positive, it is the next direction
in its module's prefix ($j=k_m+1$), and its cost fits the remaining budget.
Ties use stable adapter/module identifiers followed by direction index.
For $D=\sum_m(r_m-\ell_m)$ candidates, sorting takes $O(D\log D)$ time and
selection scans the list once. Unequal direction costs make this a greedy
heuristic for the integer objective, potentially leaving a small budget
remainder. Its output is a retained-rank manifest with the exact compact cost
$\sum_m c_m k_m$. Quality and serving use $\ell_m=\max(1,\lceil u_m/2\rceil)$;
the no-floor quality ablation uses $\ell_m=1$. Both policies share the
ordered-factor cache and runtime metadata.

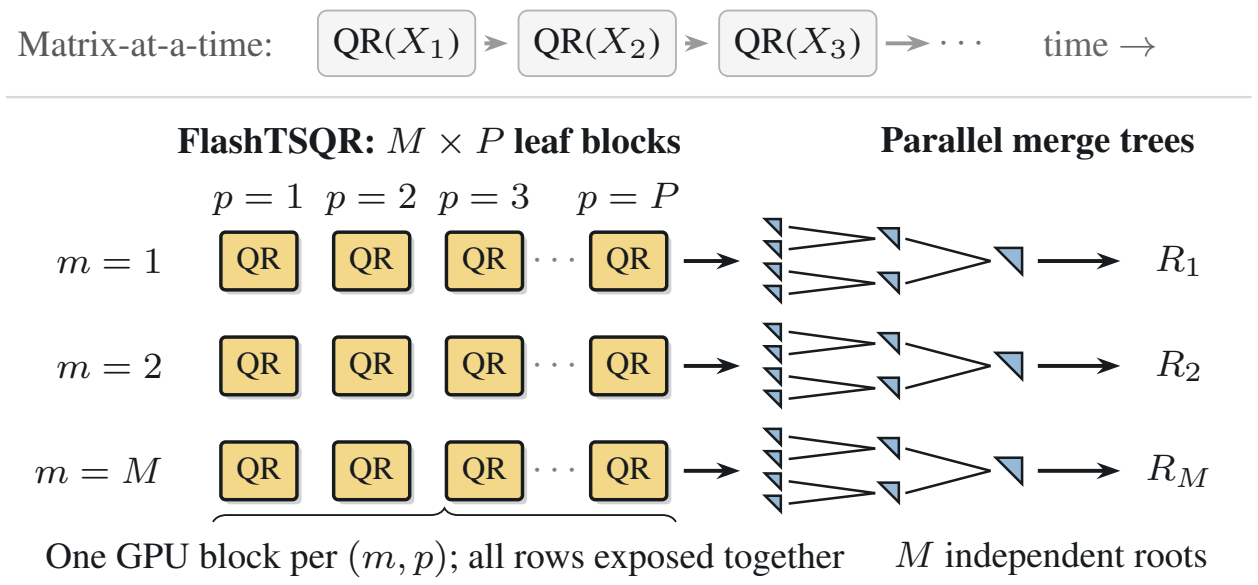
\begin{figure}[t]
  \centering
  \resizebox{\linewidth}{!}{\begingroup%
\definecolor{ftblue}{HTML}{7CA8CE}%
\definecolor{ftgreen}{HTML}{9BBD7E}%
\definecolor{ftgold}{HTML}{F1CE6C}%
\definecolor{ftink}{HTML}{171B1E}%
\definecolor{ftneutral}{HTML}{E5E7E9}%
\colorlet{ftteal}{ftgreen}%
\colorlet{ftnavy}{ftink}%
\begin{tikzpicture}[
  x=1cm,y=1cm,text=ftink,font=\fontsize{8}{9.5}\selectfont,
  >={Stealth[length=1.8mm,width=1.25mm]},
  flow/.style={->,draw=ftink,line width=.85pt},
  replay/.style={->,draw=ftink,line width=.85pt},
  lab/.style={font=\fontsize{8}{9.5}\selectfont,align=center,inner sep=1pt},
  head/.style={lab,font=\fontsize{8}{9.5}\selectfont\bfseries},
  op/.style={drop shadow={shadow xshift=1pt,shadow yshift=-1pt,
    fill=black,opacity=.18},draw=ftink,line width=.75pt,
    rounded corners=2.5pt,fill=ftgold!80,text=ftink,
    minimum width=9mm,minimum height=10mm,inner sep=2pt,align=center},
  tri/.style={draw=ftink,fill=ftblue!80,line width=.55pt},
  smallmat/.style={draw=ftink,fill=ftgreen!80,line width=.55pt},
]

\newcommand{\ftmatrix}[6]{%
 \foreach \layer in {#5,...,0}{
  \begin{scope}[shift={(#1+.055*\layer,#2+.075*\layer)}]
   \filldraw[drop shadow={shadow xshift=1pt,shadow yshift=-1pt,
      fill=black,opacity=.18},draw=ftink,fill=#6!18,line width=.75pt]
      (0,0) rectangle (#3,#4);
   \foreach \row in {0,...,4}{
    \foreach \col in {0,1}{
     \pgfmathtruncatemacro{\shade}{60+10*mod(\row+\col,2)}
     \filldraw[fill=#6!\shade!white,draw=ftink!40,line width=.25pt]
       (\col*#3/2,\row*#4/5) rectangle ++(#3/2,#4/5);
    }
   }
   \draw[ftink,line width=.75pt] (0,0) rectangle (#3,#4);
  \end{scope}
 }
}
\newcommand{\fttriangle}[3]{%
 \filldraw[tri] (#1,#2+#3)--(#1+#3,#2+#3)--(#1+#3,#2)--cycle;
}

\node[lab,anchor=west,text=black!65] at (.30,4.11) {Matrix-at-a-time:};
\foreach \xx/\idx in {3.23/1,4.76/2,6.29/3}{
 \node[lab,draw=black!30,fill=black!4,rounded corners=2pt,
       minimum width=12mm,minimum height=5mm] at (\xx,4.11) {QR($X_{\idx}$)};
}
\draw[flow,draw=black!40] (3.90,4.11)--(4.08,4.11);
\draw[flow,draw=black!40] (5.43,4.11)--(5.61,4.11);
\draw[flow,draw=black!40] (6.96,4.11)--(7.30,4.11);
\node[lab,text=black!55] at (7.56,4.11) {$\cdots$};
\node[lab,anchor=west,text=black!60] at (8.12,4.11) {time $\rightarrow$};
\draw[black!15,line width=.5pt] (.25,3.70)--(9.75,3.70);
\node[head,text=ftnavy] at (3.48,3.35) {FlashTSQR: $M\times P$ leaf blocks};
\node[head,text=ftnavy] at (8.12,3.35) {Parallel merge trees};
\foreach \xx/\pp in {2.18/1,3.04/2,3.90/3,5.00/P}{
 \node[lab] at (\xx,2.92) {$p=\pp$};
}
\foreach \yy/\mm/\cc in {2.45/1/ftblue,1.65/2/ftteal,.85/M/ftgreen}{
 \node[lab,anchor=east] at (1.49,\yy) {$m=\mm$};
 \foreach \xx/\pp in {2.18/1,3.04/2,3.90/3,5.00/P}{
  \filldraw[draw=ftink,fill=ftgold!80,line width=.75pt,rounded corners=1pt,
    drop shadow={shadow xshift=1pt,shadow yshift=-1pt,fill=black,opacity=.18}]
    (\xx-.29,\yy-.22) rectangle ++(.58,.44);
  \node[lab,font=\fontsize{7}{8}\selectfont] at (\xx,\yy) {QR};
 }
 \node[lab,text=black!50] at (4.46,\yy) {$\cdots$};
 \draw[flow] (5.42,\yy)--(5.85,\yy);
 \foreach \dy in {-.25,-.08,.09,.26}{\fttriangle{6.04}{\yy+\dy-.06}{.12}}
 \foreach \dy/\parent in {-.25/-.17,-.08/-.17,.09/.17,.26/.17}{
  \draw[draw=ftink,line width=.55pt] (6.22,\yy+\dy)--(6.88,\yy+\parent);
 }
 \fttriangle{6.91}{\yy-.24}{.15}
 \fttriangle{6.91}{\yy+.10}{.15}
 \draw[draw=ftink,line width=.55pt] (7.11,\yy-.17)--(7.76,\yy);
 \draw[draw=ftink,line width=.55pt] (7.11,\yy+.17)--(7.76,\yy);
 \fttriangle{7.80}{\yy-.10}{.20}
 \draw[flow] (8.12,\yy)--(8.75,\yy);
 \node[lab] at (9.20,\yy) {$R_{\mm}$};
}
\draw[decorate,decoration={brace,amplitude=3pt},draw=ftnavy]
 (1.83,.46)--(5.36,.46);
\node[lab,text=ftnavy] at (3.60,.18) {One GPU block per $(m,p)$; all rows exposed together};
\node[lab,text=ftnavy] at (8.22,.18) {$M$ independent roots};

\end{tikzpicture}%
\endgroup%
}
  \caption{\textbf{FlashTSQR's parallel QR schedule.}
  Our batched schedule launches $M\times P$ leaf blocks across $M$ matrices and
  $P$ row tiles, then performs level-wise reductions for all $M$ merge trees. }
  \label{fig:flashtsqr-parallelism}
\end{figure}

\subsection{FlashTSQR: A GPU Primitive for Fleet Compression}
\label{sec:flashtsqr}

We develop FlashTSQR, a GPU factorization primitive tailored to the thousands
of tall, narrow factors in a LoRA fleet. Its module--tile schedule exposes
parallelism across factors and row tiles, while packed triangular merges
and implicit basis application reduce intermediate storage.

The factor construction builds on FraQ's one-sided coordinate-space
formulation~\citep{li2026fraq} and Householder tall-skinny QR
(TSQR)~\citep{demmel2012tsqr}.
For $\Delta\mat W=s\mat B\mat A$, choose
$(\mat X,\mat Y)=(\mat B,\mat A)$ if
$d_{\mathrm{out}}\le d_{\mathrm{in}}$, and
$(\mat A^\top,\mat B^\top)$ otherwise.
Thin QR of the smaller-ambient factor gives $\mat X=\mat Q\mat R$.
We form $\mat H=\mat R\mat Y$ and diagonalize the $r\times r$ core
$\mat H\mat H^\top=\mat V\mat\Sigma^2\mat V^\top$ in descending
spectral order. The balanced factors are
\begin{equation}
\begin{aligned}
  \mat F &= \mat Q(\mat V\mat\Sigma^{1/2}),\\
  \mat T &= \mat\Sigma^{\dagger/2}\mat V^\top\mat H,
\end{aligned}
\label{eq:flash-reconstruction}
\end{equation}
where $\mat\Sigma^{\dagger/2}$ is the pseudoinverse of
$\mat\Sigma^{1/2}$. We return
$(\overline{\mat B},\overline{\mat A})=(\mat F,\mat T)$ for
$\mat X=\mat B$, or $(\mat T^\top,\mat F^\top)$ otherwise.
Their first $k$ directions reconstruct the rank-$k$ truncated SVD in exact
arithmetic, with the original scale $s$ preserved on export. This avoids
forming the dense update or computing a second QR.

We map this construction to the GPU with a module--tile schedule
(Figure~\ref{fig:flashtsqr-parallelism}), grouping modules with the same
$(d_{\mathrm{in}},d_{\mathrm{out}},r)$ into batches.
For $M$ factors, each split into $P$ row tiles, a leaf launch
exposes $MP$ matrix--tile blocks; each subsequent launch merges one tree level
across all $M$ matrices. Our merge kernel exploits the two inputs'
triangular structure, storing $r(r+1)$ packed floats instead of a dense
$2r^2$ tile. Intermediate triangular factors remain packed, and saved
reflectors support implicit-$\mat Q$ application.

One implicit-$\mat Q$ application reconstructs all $r$ ordered directions.
Both policies and different budgets select prefixes from this factor cache,
avoiding repeated factorization or successive compression of already
truncated weights. Construction cost therefore depends on the original rank,
while retained rank determines exported factor sizes and runtime work.
Appendix~\ref{app:factor-construction} details precision, eigensolver dispatch,
wide-rank fallbacks, and validation; Section~\ref{sec:conversion-latency}
measures full-fleet recompression time with matched GPU backends.

\section{Rank-Specialized Execution}
\label{sec:serving}

LoRAForge's shared backend specializes the shrink and expand operations
(Section~\ref{sec:background}) for ranks exported by either allocation policy. Generic segmented kernels can still compute at padded capacity
or use poorly matched matrix-product tiles. We preserve the exported ranks
(Section~\ref{sec:retained-rank}), expose them to computation and factor access,
and specialize shrink and expand for this regime. A dispatch policy composes
these kernels with the native segmented path according to rank, shape, and
serving phase.

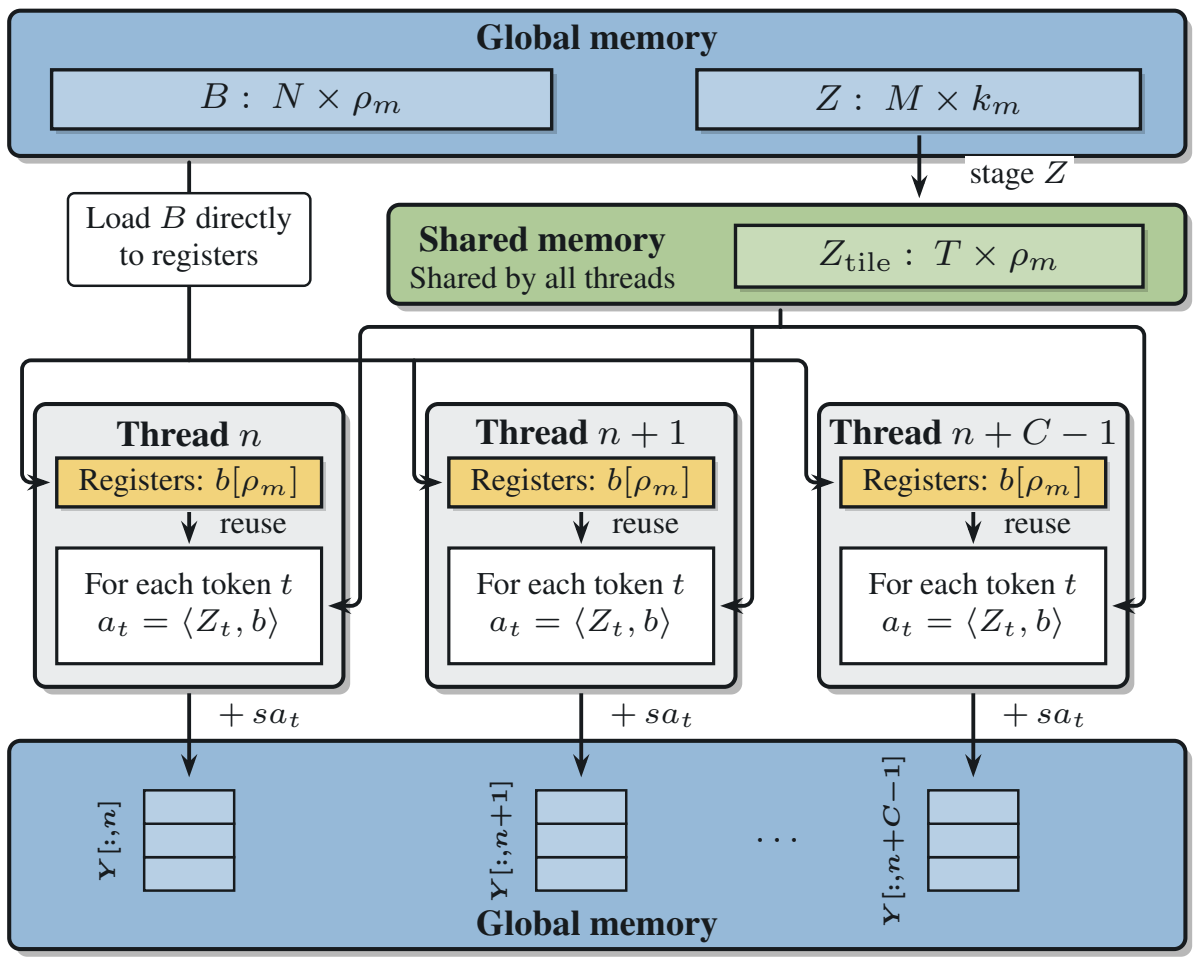
\begin{figure}[t]
\centering
\resizebox{\linewidth}{!}{\begingroup
\definecolor{exblue}{HTML}{7CA8CE}
\definecolor{exgreen}{HTML}{9BBD7E}
\definecolor{exgold}{HTML}{F1CE6C}
\definecolor{exink}{HTML}{171B1E}
\definecolor{exneutral}{HTML}{E5E7E9}
\begin{tikzpicture}[x=1cm,y=1cm,text=exink,font=\fontsize{8}{9.5}\selectfont,
 >={Stealth[length=1.8mm,width=1.25mm]},
 title/.style={font=\fontsize{10}{12}\selectfont\bfseries,anchor=west},
 label/.style={align=center,inner sep=1pt},
 small/.style={label,font=\fontsize{7}{8.2}\selectfont},
 tinylabel/.style={label,font=\fontsize{7}{8.2}\selectfont},
 head/.style={label,font=\fontsize{8}{9.5}\selectfont\bfseries},
 card/.style={draw=exink,fill=#1!80,line width=.9pt,rounded corners=2.5pt,
 drop shadow={shadow xshift=1.5pt,shadow yshift=-1.7pt,fill=black,opacity=.28}},
 arrow/.style={->,draw=exink,line width=.85pt},
 cell/.style={draw=exink,line width=.55pt,fill=#1!80}]
\draw[card=exblue] (8.7,6.15) rectangle (17.85,7.28);
\node[head] at (13.275,7.05) {Global memory};
\draw[draw=exink,fill=exblue!55,line width=.75pt,drop shadow={shadow xshift=1pt,shadow yshift=-1pt,fill=black,opacity=.18}] (9.03,6.36) rectangle (12.92,6.82);
\node[label] at (10.975,6.59) {$B:\ N\times \rho_m$};
\draw[draw=exink,fill=exblue!55,line width=.75pt,drop shadow={shadow xshift=1pt,shadow yshift=-1pt,fill=black,opacity=.18}] (14.05,6.36) rectangle (17.5,6.82);
\node[label] at (15.775,6.59) {$Z:\ M\times k_m$};
\draw[card=exgreen] (11.65,5.0) rectangle (17.85,5.77);
\node[head] at (12.85,5.48) {Shared memory};
\node[small] at (12.85,5.18) {Shared by all threads};
\draw[draw=exink,fill=exgreen!55,line width=.75pt,drop shadow={shadow xshift=1pt,shadow yshift=-1pt,fill=black,opacity=.18}] (14.35,5.13) rectangle (17.52,5.61);
\node[label] at (15.935,5.37) {$Z_{\mathrm{tile}}:\ T\times \rho_m$};
\draw[arrow=exgreen] (15.78,6.31) -- (15.78,5.82);
\node[tinylabel,fill=white,text=exink] at (16.56,6.0) {stage $Z$};
\draw[draw=exink,line width=.85pt,line cap=round] (10.1,6.1) -- (10.1,4.56);
\node[small,text=exink,fill=white,draw=exink,line width=.55pt,
 rounded corners=1.5pt,inner xsep=4pt,inner ysep=3pt]
 at (10.1,5.5) {Load $B$ directly\\to registers};
\draw[draw=exink,line width=.85pt,line cap=round] (14.7,4.95) -- (14.7,4.82);
\draw[card=exblue] (8.7,-.02) rectangle (17.85,1.6);
\node[head] at (13.275,.16) {Global memory};
\foreach \x/\n/\yn in {8.9/n/n,11.95/{n+1}/{n\mathord{+}1},15.0/{n+C-1}/{n\mathord{+}C\mathord{-}1}}{
 \draw[card=exneutral] (\x,2.02) rectangle (\x+2.4,4.22);
 \node[head] at (\x+1.2,3.99) {Thread $\n$};
 \draw[draw=exink,fill=exgold!90,line width=.75pt,drop shadow={shadow xshift=1pt,shadow yshift=-1pt,fill=black,opacity=.18}] (\x+.17,3.42) rectangle (\x+2.23,3.8);
 \node[small] at (\x+1.2,3.61) {Registers: $b[\rho_m]$};
 \draw[arrow=exblue] (\x+1.2,3.38) -- (\x+1.2,3.13);
 \node[tinylabel,text=exink,anchor=west] at (\x+1.4,3.27) {reuse};
 \draw[draw=exink,fill=white,line width=.7pt] (\x+.17,2.2) rectangle (\x+2.23,3.1);
 \node[tinylabel] at (\x+1.2,2.65) {For each token $t$\\$a_t=\langle Z_t,b\rangle$};
 \draw[arrow=exink] (\x+1.2,1.97) -- (\x+1.2,1.315);
 \node[tinylabel,text=exink,anchor=west] at (\x+1.38,1.8) {$+\,s a_t$};
 \foreach \j in {0,...,2}{\draw[cell=exblue,fill=exblue!55] (\x+.85,.435+\j*.26) rectangle ++(.7,.26);}
 \node[small,rotate=90,scale=.85,font=\fontsize{6}{7}\selectfont\boldmath] at (\x+.58,.825) {$Y[\mathord{:},\!\yn]$};
}
\draw[arrow=exgreen,rounded corners=2pt,line cap=round,line join=round]
 (14.7,4.82) -- (11.43,4.82) -- (11.43,2.65) -- (11.18,2.65);
\draw[arrow=exgreen,rounded corners=2pt,line cap=round,line join=round]
 (14.7,4.82) -- (17.53,4.82) -- (17.53,2.65) -- (17.28,2.65);
\draw[arrow=exgreen,rounded corners=2pt,line cap=round,line join=round]
 (14.48,4.82) -- (14.48,2.65) -- (14.23,2.65);
\draw[arrow=exblue,rounded corners=2pt,line cap=round,line join=round]
 (10.1,4.56) -- (8.8,4.56) -- (8.8,3.61) -- (9.02,3.61);
\draw[arrow=exblue,rounded corners=2pt,line cap=round,line join=round]
 (10.1,4.56) -- (14.9,4.56) -- (14.9,3.61) -- (15.12,3.61);
\draw[arrow=exblue,rounded corners=2pt,line cap=round,line join=round]
 (11.85,4.56) -- (11.85,3.61) -- (12.07,3.61);
\node at (14.7,.825) {$\cdots$};
\end{tikzpicture}
\endgroup}
\caption{\textbf{RowExpand hardware mapping.}
Each thread loads one $B$ row directly into registers and reuses it across $T$ tokens. The shared $Z$ tile
serves $C$ output threads, each completing its rank reduction locally and adding $s a_t$ to the corresponding output. Here $M$ and $N$ denote input tokens and output channels, $k_m$ is the retained rank, and
$\rho_m\geq k_m$ is the padded rank bucket width.
}
\label{fig:hip-expand-design}
\end{figure}

\paragraph{Rank-aware shrink.}
In shrink, retained rank is the output width: each direction computes
$Z[t,j]=\langle X[t,:],A[j,:]\rangle$ over the full input dimension
$d_{\mathrm{in}}$. Compression reduces the number of these dot products,
so executing all rows of a padded $A$ buffer can erase its computational
benefit. We expose the effective rank to both direct reduction and segmented
matrix-product paths.

For short decode batches, our direct kernel assigns one cooperating thread
group to a token and an output direction. A device-side token-to-adapter map
selects the weights and rank; groups beyond that rank return before reading
operands. Threads vector-load contiguous activation and weight elements,
accumulate products in FP32, and combine their partial sums through shuffle
operations. One thread writes the resulting $Z[t,j]$, completing the input
reduction without a global partial-sum buffer or a second reduction kernel.
Fused projections pack their $A$ slices using the largest retained slice
rank as a common bound, while physical adapter-slot and output-row strides retain the pool
capacity; shorter slices remain zero-padded. Thus, effective rank controls
which directions execute without requiring tightly allocated buffers.

For larger token batches, we retain segmented matrix-product tiling to reuse
operands across tokens. Our rank-aware variant exits output-column blocks
wholly beyond the effective rank before the input-dimension reduction loop;
partially live blocks retain their masks. This avoids executing matrix
products on entirely padded columns while preserving the tiled path for
prefill. Both paths pass live $Z$ entries to subsequent expand in the same layout;
dispatch selects between them by phase and shape.
Appendix~\ref{app:execution-implementation} details the platform mappings
and fused layouts.

\paragraph{RowExpand: thread-owned weight rows with cross-token reuse.}
Conventional segmented expand kernels organize work as matrix-product tiles.
For example, SGLang's CSGMV computes a token-by-output tile per program,
loading $Z$ and $B$ tiles and accumulating their products over the rank
dimension in the implementation we examine (Appendix~\ref{app:execution-implementation}). This already reuses weights within a token
tile, but small ranks and short token chunks can underfill its matrix-product
tiles. RowExpand uses a thread-local reduction designed for this regime
(Figure~\ref{fig:hip-expand-design}).
It stages the small intermediate $Z$ tile in shared memory.
Each thread owns one output channel $n$, loads the corresponding compact
weight row $B[n,:]$ into registers, and reuses it across $T$ token rows of
the same adapter segment. The shared $Z$ tile serves all output threads
in the block. For each token, the owning thread completes the rank reduction
locally and adds the scaled result to the base output, without a cross-thread
reduction. Weight rows are thus reused across tokens, while activations are
reused across output channels. The row in RowExpand refers to a weight row
of $B$, corresponding to a column of the output.

Compact row access keeps the register working set tied to the retained rank.
We map nearby ranks to compile-time bucket widths $\rho_m\geq k_m$,
zero-padding unused entries while preserving which directions are live.
The bucket specializes weight loads and the local reduction without making
an adapter's maximum checkpoint rank the width of every module. Small
retained ranks make this register-resident mapping practical; dispatch
retains the native path when ranks or shapes make it unfavorable.

\paragraph{Rank-, shape-, and phase-aware composition.}
A narrow rank alone does not determine the best execution path: token count,
projection shape, and serving phase also affect reuse and parallelism.
We therefore compose the custom kernels with the rank-aware native path
using a dispatch policy frozen before evaluation. Shrink and expand can
select different paths, and prefill can differ from decode. Unsupported or
unfavorable cases use the native kernels. Dispatch changes how the exported
factors are executed, without changing their retained directions or update.

\section{Evaluation}
\label{sec:experiments}

We evaluate shared construction and execution, then allocation tradeoffs.
Paired controls test the runtime under both policies; deployment curves use
frozen spectral exports. Quality sweeps and serving share adapter cohorts and, at $1/12$,
the same frozen rank-floored spectral exports.

\subsection{Setup}

\paragraph{Adapter fleets.}
Spectral characterization (Figure~\ref{fig:rank-utilization}), quality
evaluation, and serving use the same 35-adapter Llama and 50-adapter Qwen
cohorts, based on Llama~3.1~\citep{grattafiori2024llama3} and
Qwen3~\citep{yang2025qwen3}, respectively. The Qwen cohort retains the full-data
HypergraphFormer checkpoint and removes its three smaller-data variants by
checkpoint identity, rather than quality or speed scores. Its spectral
allocation is recomputed from original factors.

\paragraph{Policies, references, and budgets.}
Quality sweeps compare \lfuniform{} and \lfspectral{} with external CTS,
using Base and Original as references. Divisors $d\in\{12,8,4,2\}$ set
\lfuniform{}'s compact cost and the shared ceiling. The main
\lfspectral{} quality configuration uses the rank floor in
Section~\ref{sec:allocation}; its ablation is labeled w/o floor.
The Qwen execution control pairs both policies at $1/12$, using the same
rank-floored spectral exports as quality evaluation. AMD and GH200 deployment
sweeps also use these frozen $1/12$ \lfspectral{} exports. CTS uses the authors' JD-Full implementation
\citep{brueelgabrielsson2024compressthenserve}, counting each shared basis once
and all adapter cores against the same ceiling. Fits are selected by
reconstruction objective and evaluated for quality only
(Appendix~\ref{app:cts-quality}).

\paragraph{Serving configuration.}
We co-locate two independent SGLang engines: Qwen3-4B-Instruct-2507 with
50 adapters and Llama-3.1-8B-Instruct with 35, on one MI250X GCD (64\,GB)
or one GH200 GPU. LoRAForge extends SGLang's segmented LoRA
path~\citep{zheng2024sglang}. Its kernels support HIP on AMD and CUDA on NVIDIA,
with BF16 operands and FP32 accumulation. Both platforms use BF16, tensor
parallelism one, 16,384 KV tokens per model, and Qwen/Llama adapter-storage
budgets of 14/10\,GiB, including
custom shadow weights. Each of five seeds sends 1,700 requests, 20 per adapter,
in a mixed Poisson arrival stream. Requests, routing, and arrivals are matched
across arms within each platform. Appendix~\ref{app:execution-implementation}
records platform mappings, frozen dispatch settings, and graph/correctness checks;
Appendix~\ref{app:gh200-mixed-serving} gives GH200 residency details.

\paragraph{Metrics.}
Quality is summarized by adapter-macro averages of task-native scores
(accuracy, micro-F1, chrF, or structural metrics).
Request and output-token throughput divide completed requests and actual
generated tokens, respectively, by the same wall-clock measurement interval.
TTFT measures time to first emitted text; request-level TPOT is
$(T_{\mathrm{end}}-T_{\mathrm{first}})/(n_{\mathrm{out}}-1)$ for outputs of at
least two tokens. TPOT is computed per request, without pooling inter-token
intervals. We report TTFT P50 and TPOT P99 alongside both throughput
metrics and output lengths. Short outputs can make TPOT tails sensitive to
one scheduler stall. Quantiles are computed within each run and averaged
across seeds; shaded bands show sample standard deviations, not confidence
intervals.

\paragraph{Work controls.}
For serving, Base Only disables LoRA and replays exact per-request output-token
counts with EOS ignored, providing a no-LoRA compute reference.
Single-fleet output-trace versions are specified in
Appendix~\ref{app:single-fleet-serving}.
Separate fixed-output backend ablations use identical token counts for both
compared backends. The six-cell control in Section~\ref{sec:backend-controls}
additionally matches the stopping contract and fixes KV capacity across all
allocations and paths. Matching output counts alone does not ensure equal
scheduler admission: output caps and EOS also affect KV reservation.
Appendix~\ref{app:mixed-tpot} examines this distinction.

\subsection{Full Task Quality Across Four Budgets}
\label{sec:quality-budgets}

\begin{table}[t]
\centering\fontsize{8.5}{10}\selectfont
\caption{\textbf{Task quality under matched compact-parameter ceilings.}
Adapter-macro native scores (0--100, $\uparrow$) over 21,592 Qwen and 26,336
Llama requests per arm. Shaded rows give Base and Original references;
bold marks the best compressed score per budget. \lfuniform{} uses
a uniform retention ratio; \lfspectral{} uses spectral allocation with
rank floors. Both are LoRAForge configurations. CTS uses JD-Full.}
\label{tab:quality-fleets}
\setlength{\tabcolsep}{0pt}
\renewcommand{\arraystretch}{1.15}
\begin{tabular}{@{}p{90pt}*{4}{>{\raggedleft\arraybackslash}p{\dimexpr(\linewidth-90pt)/4\relax}}@{}}
\toprule
 & \multicolumn{4}{c}{Retained parameter budget} \\
\cmidrule(l){2-5}
Configuration & $1/2$ & $1/4$ & $1/8$ & $1/12$ \\
\midrule
\rowcolor{black!6}\multicolumn{5}{>{\columncolor{black!6}[0pt][0pt]}p{\linewidth}@{}}{\textbf{Qwen50}\hfill{\fontsize{8}{9.5}\selectfont Base: 43.12\quad Original: 70.21}} \\
\addlinespace[2pt]
CTS & 70.02 & 69.44 & 67.90 & 62.54 \\
\textcolor{navy}{\textbf{\lfuniform{}}} & 69.99 & 69.67 & 69.09 & 68.33 \\
\textcolor{navy}{\textbf{\lfspectral{}}} & \textbf{70.22} & \textbf{70.23} & \textbf{70.11} & \textbf{69.41} \\
\addlinespace[3pt]
\midrule
\rowcolor{black!6}\multicolumn{5}{>{\columncolor{black!6}[0pt][0pt]}p{\linewidth}@{}}{\textbf{Llama35}\hfill{\fontsize{8}{9.5}\selectfont Base: 52.20\quad Original: 71.13}} \\
\addlinespace[2pt]
CTS & 69.64 & 69.11 & 67.99 & 67.21 \\
\textcolor{navy}{\textbf{\lfuniform{}}} & \textbf{71.10} & 70.71 & 70.04 & 68.68 \\
\textcolor{navy}{\textbf{\lfspectral{}}} & 71.00 & \textbf{70.91} & \textbf{70.44} & \textbf{69.93} \\
\bottomrule
\end{tabular}

\end{table}

Table~\ref{tab:quality-fleets} compares CTS with both LoRAForge policies
at four matched compact-parameter ceilings ($1/2$, $1/4$, $1/8$, and $1/12$),
with Base and uncompressed Original as references. We use all 21,592 frozen
Qwen validation requests and 26,336
Llama requests (24,536 classification and 1,800 reviewed translation).
Input token IDs, task caps, and greedy natural stopping are matched across
arms; LoRA exports use BF16. Llama Base is a GH200 BF16 eager control;
the other Llama arms run on MI250X with the same inputs and scoring.
Scores are descriptive adapter-level means of task-native metrics on a
0--100 scale, rather than pooled accuracy. These are previously exposed
public evaluation pools.

Both LoRAForge policies retain high fleet-average scores under tight budgets.
\lfuniform{} exceeds CTS in seven of eight fleet--budget settings;
Qwen at $1/2$ is the exception (69.99 versus 70.02).
\lfspectral{} improves on \lfuniform{} in seven settings, with gains of
1.08 and 1.25 points on Qwen and Llama at $1/12$. Its corresponding drops
from Original are 0.80 and 1.20 points, versus CTS's 7.67 and 3.92 points.
Uniform retention remains slightly higher on Llama at $1/2$ (71.10 versus
71.00). Spectral allocation therefore offers additional quality retention
in most measured settings, rather than defining the system's only useful
configuration. These scores do not imply lower physical memory or faster
serving. Appendix~\ref{app:quality-regressions} reports per-adapter regressions.

\subsection{Full-Fleet Recompression with Matched Backends}
\label{sec:conversion-latency}
\label{sec:conversion-gh200}

We measure the factor-construction stage of full-fleet LoRA recompression
on one GH200 using local implementations of FLoRIST~\citep{ramesh2026florist},
SpecTraL~\citep{ramesh2026spectral}, and FraQ~\citep{li2026fraq}.
Our primary control compares
FraQ with and without FlashTSQR under identical precision, batch schedules,
and GPU eigensolver; only QR changes. Each timed pass constructs full ordered
factors for later rank selection, excluding allocation, checkpoint export,
and serving. All methods pass the same reconstruction and spectrum checks.

\begin{figure}[t]
\centering
\includegraphics[width=\linewidth]{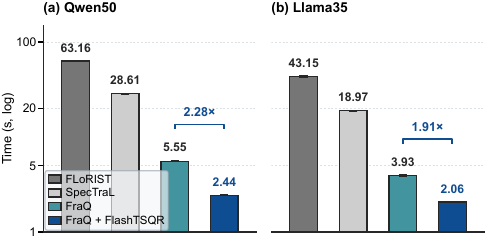}
\caption{\textbf{Full-fleet recompression time on one GH200.}
FLoRIST and SpecTraL process one matrix pair per call. Both FraQ arms use
the same precision, batching, and GPU FP64 eigensolver; only QR changes.
Bars show median factor-construction times with min--max whiskers
over five runs; brackets compare FraQ medians.}
\label{fig:conversion-gh200}
\end{figure}

Figure~\ref{fig:conversion-gh200} shows that FlashTSQR reduces recompression time
from 5.55 to 2.44\,s on Qwen50 and from 3.93 to 2.06\,s on Llama35,
yielding $2.28\times$ and $1.91\times$ whole-pipeline speedups over matched
FraQ. Values are medians of five full-fleet runs. FLoRIST and SpecTraL provide
broader comparisons with different factorization and batching strategies.
Appendix~\ref{app:gh200-conversion-protocol} gives implementation settings,
timing boundaries, and numerical acceptance criteria.
The supplementary MI250X study (Appendix~\ref{app:mi250x-conversion})
includes backend-dependent solvers and fallbacks, so it does not isolate
hardware differences.

\subsection{End-to-End Serving Performance}
\label{sec:exp-serving}
\label{sec:mixed-serving}

\begin{figure*}[t]
\centering
\begin{subfigure}{\textwidth}
\centering
\includegraphics[width=\linewidth]{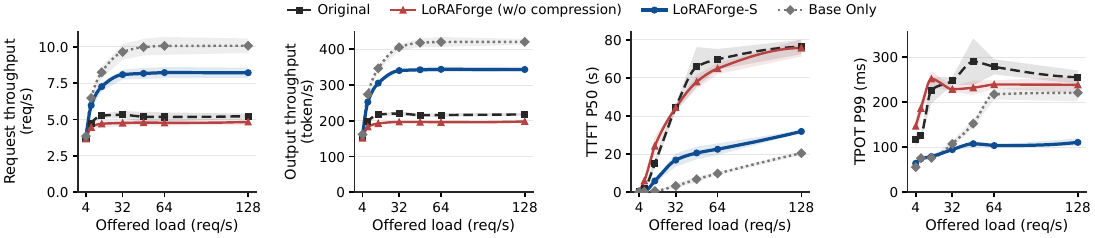}
\caption{\textbf{AMD MI250X}}
\label{fig:mixed-serving}
\end{subfigure}
\par\medskip
\begin{subfigure}{\textwidth}
\centering
\includegraphics[width=\linewidth]{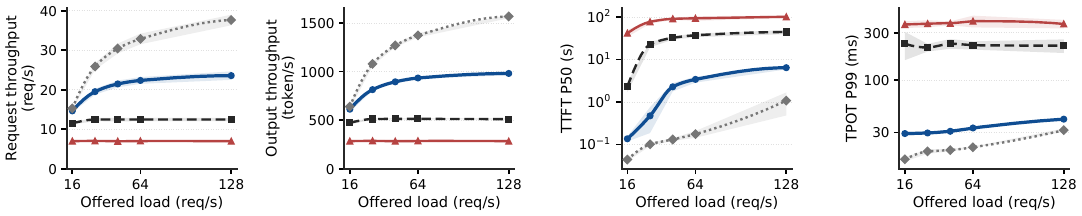}
\caption{\textbf{NVIDIA GH200}}
\label{fig:mixed85-gh200}
\end{subfigure}
\caption{\textbf{Single-device serving of two models and 85 adapters on AMD and NVIDIA.}
Each row co-locates Qwen50 and Llama35. Markers show five-seed means and bands
sample SD; smooth lines guide the eye between measured loads.
LoRA arms use natural stopping; Base Only replays \lfspectral{}'s per-request
output lengths. Both platforms use the same rank-floored $1/12$ spectral exports; compare configurations
within each row. GH200 latency axes are logarithmic.}
\label{fig:mixed-platforms}
\end{figure*}

Figure~\ref{fig:mixed-platforms} compares four configurations on co-located
Qwen50 and Llama35: Original uses native SGLang kernels; LoRAForge (w/o
compression) applies our rank-aware path to original adapters; \lfspectral{} uses frozen spectral exports; Base Only disables LoRA and replays \lfspectral{}'s output
lengths. LoRA arms retain natural stopping. \lfspectral{} improves token throughput
over Original by $1.59\times$ on MI250X at 64 offered requests/s and
$1.92\times$ on GH200 at 128 requests/s, while reducing GH200 TPOT P99 from
222.4 to 40.8\,ms. The optimized path alone yields no comparable gains,
highlighting the benefit of pairing compression with execution suited to
retained ranks. Compression also eliminates timed adapter loads under the
same storage ceilings, versus roughly 840 loads per Original run at these
rates. The gains therefore combine residency, execution, and output-length
effects; a matched-output MI250X control still retains $1.48\times$
(Appendix~\ref{app:fixedwork64}). Section~\ref{sec:backend-controls} examines
these contributions separately. The benefit also holds in a vLLM~\citep{kwon2023vllm} implementation: with a common four-decode-step scheduling policy, \lfspectral{} delivers $1.50$--$1.54\times$ the token throughput
of Original at 32--128 offered requests/s across five seeds
(Appendix~\ref{app:vllm-serving}, Figure~\ref{fig:mixed85-vllm}).

Comparisons in Figure~\ref{fig:mixed-platforms} are within platforms, using
the same rank-floored $1/12$ spectral exports across platforms. Throughput includes queue drain
over 1,700 requests and does not
establish sustainable capacity at the offered rate. Base Only's higher
MI250X TPOT P99 likewise does not establish a decode advantage for \lfspectral{}:
exact output caps and ignored EOS change KV reservation and admission.
Applying this stopping contract to unchanged \lfspectral{} raises TPOT P99 in
the single-seed diagnostic (Appendix~\ref{app:mixed-tpot}). Independent
single-fleet results appear in Appendix~\ref{app:single-fleet-serving}.

\subsection{Shared Runtime and Policy Tradeoffs}
\label{sec:backend-controls}

\begin{figure}[!t]
\centering
\includegraphics[width=\linewidth]{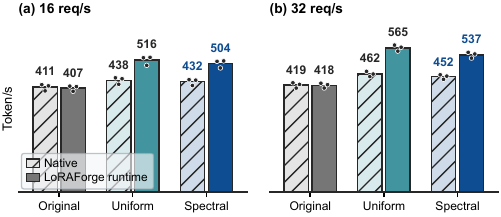}
\caption{\textbf{The same runtime benefits both compressed policies.}
Qwen50 at (a) 16 and (b) 32 requests/s, with matched inputs, output lengths,
stopping rules, and capacity limits. Bars show mean token throughput and dots
three input seeds. %
}
\label{fig:allocation-execution}
\end{figure}

Figure~\ref{fig:allocation-execution} crosses Original, Uniform, and Spectral
weights with Native and LoRAForge execution on Qwen50; compressed arms use $1/12$.
Three seeds at 16/32 offered requests/s share inputs, arrivals, output lengths,
stopping rules, KV capacity, and resident-adapter limits. Spectral uses the
same rank-floored $1/12$ exports as quality evaluation. The shared runtime adds $1.18\times$/$1.22\times$ for
Uniform and $1.17\times$/$1.19\times$ for Spectral, versus
$0.99\times$/$1.00\times$ for Original. Thus execution gains extend to
both policies, rather than depending on spectral reallocation.
Spectral/native reaches $1.05\times$/$1.08\times$ Original/native throughput;
adding the shared runtime raises these ratios to $1.23\times$/$1.28\times$.
With this runtime, \lfspectral{} is 2.5\%/4.9\% slower than \lfuniform{},
but scores 69.41 versus 68.33. Its engine-lifetime allocated-memory peak is
33.88 versus 24.60\,GiB, including loading, validation, and graph capture.
These matched-budget configurations expose a quality--resource tradeoff,
not a uniformly best allocator. Throughput ratios are paired geometric means;
Appendix~\ref{app:factorial} records timing and memory boundaries.

\begin{figure}[!t]
\centering
\includegraphics[width=\linewidth]{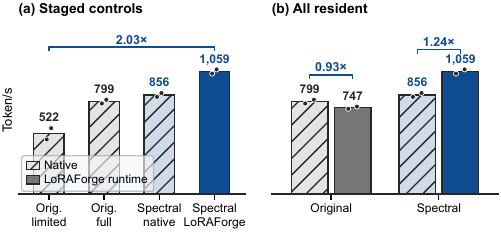}
\caption{\textbf{Capacity, compression, and execution controls.}
GH200 Mixed85 at 128 requests/s, two paired seeds with matched output work.
Bars are mean token throughput; dots are seeds and ratios divide means.
(a) Original/native limited and full, then Spectral/native and
Spectral/LoRAForge full. (b) Both weight sets fully resident; brackets compare
execution paths. Full capacity allows extra memory and has zero timed paging.
}
\label{fig:serving-attribution}
\end{figure}

Residency provides another source of gain. Figure~\ref{fig:serving-attribution}
crosses weights, execution, and capacity on GH200 at 128 requests/s, using
two paired seeds with matched output lengths and fixed KV and runtime
settings. Expanding Original/native from limited to full residency raises
throughput by $1.53\times$ with unchanged weights and execution. Full capacity
removes timed paging by granting each arm the memory it needs, unlike the
fixed-byte-budget deployment comparison. It also changes admissible adapter
diversity, so the gain measures a capacity intervention rather than DMA alone.

Execution gains persist when all adapters fit. With zero timed paging,
Spectral/native is $1.07\times$ faster than fully resident Original/native;
optimized execution adds $1.24\times$ on the same compressed weights,
for a combined $1.32\times$. Applying that path to fully resident Original
instead yields $0.93\times$. Thus residency and execution provide complementary
benefits, and the optimized path is effective specifically on the
retained-rank representation. These controls neither partition the
five-seed natural-EOS deployment gain nor isolate RowExpand: the execution
ratios include compact access, shrink, expansion, and dispatch.
Section~\ref{sec:kernel-ablations} examines the kernel mechanisms.

\subsection{Shrink, RowExpand, and Dispatch Ablations}
\label{sec:kernel-ablations}

\begin{figure}[!t]
\centering
\includegraphics[width=\linewidth]{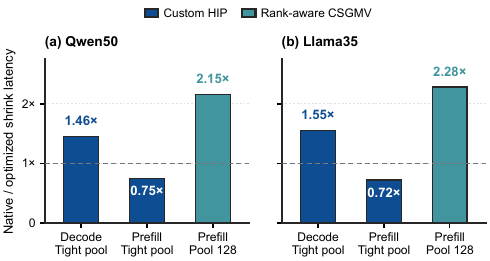}
\caption{\textbf{Shrink optimizations serve different execution regimes.}
Native/tested shrink-only latency on MI250X, summarized over ranks 1--16.
Tight pools match the retained rank; Pool 128 keeps capacity fixed at 128.
Each bar uses its own matched native baseline. Values are geometric means
of per-case median ratios, with cache flushing outside the timed interval.}
\label{fig:shrink-ablation}
\end{figure}

\paragraph{Direct reduction and effective-rank pruning.}
Figure~\ref{fig:shrink-ablation} separates shrink from expansion using the
frozen rank sweep. With tight pools, custom HIP shrink reaches
$1.46\times$/$1.55\times$ native speed on Qwen/Llama decode cases, but only
$0.75\times$/$0.72\times$ on prefill. Thus, direct reduction helps small
token batches but is not a universal replacement for tiled CSGMV.
With capacity-128 pools, rank-aware CSGMV instead reaches
$2.15\times$/$2.28\times$ native speed on prefill by skipping column blocks
beyond the effective rank. On the same prefill grid with tight pools,
rank-aware CSGMV is approximately $1.00\times$ native speed on both fleets,
consistent with the gain coming from avoiding padded work. These are
shrink-only microbenchmarks, not additive shares of end-to-end speedup;
they support selecting the reduction path by phase, shape, and rank.
Appendix~\ref{app:rank-kernels} specifies the aggregation and operands.

\begin{figure}[!t]
\centering
\includegraphics[width=\linewidth]{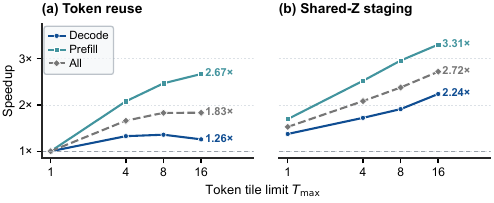}
\caption{\textbf{RowExpand's two reuse mechanisms.}
(a) Staged expand versus its $T_{\max}=1$ variant; (b) direct-Z versus staged-Z
at the same tile limit. Ratios are geometric means over 128 decode and 128
prefill cases. %
}
\label{fig:expand-ablation}
\end{figure}

\paragraph{Cross-token reuse and shared activation staging.}
We ablate RowExpand's two reuse mechanisms across 256 expand-only conditions,
varying the token-tile limit with and without shared-Z staging on matched
operands. Figure~\ref{fig:expand-ablation} shows that, with staging enabled,
increasing the limit from 1 to 16 yields a $1.83\times$ geometric-mean speedup
over all cases and $2.67\times$ on prefill. At the same tile limit of 16,
staging is $2.72\times$ faster than direct-Z access. The native comparison in Figure~\ref{fig:expand-native} shows that
together these mechanisms make RowExpand $1.24\times$ faster than native CSGMV overall and
$1.54\times$ on prefill.

Decode instead peaks at a tile limit of 8 ($1.07\times$ native) and averages
$0.99\times$ at 16. The best tile therefore depends on the workload; larger
tiles are not universally better. Staging also changes register and local-memory
use, so its gain is not solely a measure of HBM traffic reduction.
Appendix~\ref{app:expand-ablation} provides the workload grid, numerical checks,
and resource measurements.

\begin{figure}[!t]
\centering
\includegraphics[width=\linewidth]{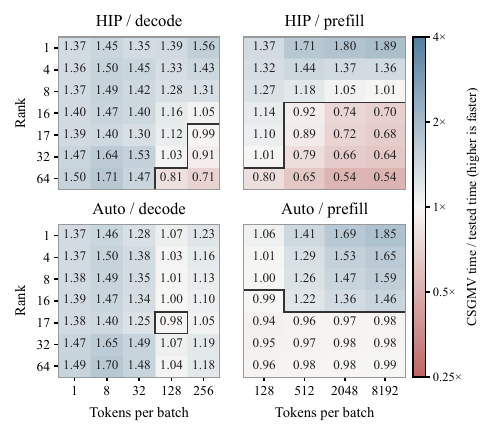}
\caption{\textbf{Rank-aware composition extends the useful kernel regimes.}
Qwen, tight rank capacity: geometric-mean native/tested shrink-plus-expand
latency over four projection shapes. HIP pairs our custom shrink with
RowExpand. %
}
\label{fig:rank-kernel}
\end{figure}

\paragraph{Why dispatch must account for rank and batch shape.}
Figure~\ref{fig:rank-kernel} compares forced custom execution with the frozen
automatic policy. At rank 64, forced custom kernels achieve $1.47\times$
native speed for 32-token decode batches but fall to $0.71\times$ at 256
tokens. For rank-16 prefill with 2,048 tokens, forcing both custom stages gives
$0.74\times$, whereas rank-aware CSGMV shrink combined with RowExpand reaches
$1.36\times$. These crossovers motivate phase-, rank-, and shape-aware
composition instead of a single rank cutoff. The policy is not retuned on
this sweep and still loses on some shapes. %

\subsection{Discussion and Limitations}
\label{sec:limitations}

The shared runtime benefits both tested compression policies, but Uniform's
execution evidence covers one Qwen budget and two offered loads.
Spectral energy does not guarantee per-adapter quality or fairness, and
the public evaluation pools are historically exposed. Neither parameter
budgets nor compact counts bound latency or physical memory.
Deployment gains combine execution, residency, and generation behavior;
matched-work controls narrow their interpretation. Two platforms support
applicability beyond one implementation, not universal portability.
Joint quality--runtime optimization and preparation-to-serving amortization
remain future work (Appendix~\ref{app:evidence-boundaries}).

\FloatBarrier
\section{Related Work}
\label{sec:related}

\paragraph{Training-time adaptation and post-hoc compression.}
AdaLoRA, DyLoRA, SoRA, and rsLoRA modify training-time ranks or scaling
\citep{zhang2023adalora,valipour2023dylora,ding2023sora,kalajdzievski2023rslora}.
QLoRA quantizes the backbone; DoRA separates magnitude and direction;
PiSSA uses spectral initialization; VeRA shares frozen random factors
\citep{dettmers2023qlora,liu2024dora,meng2024pissa,kopiczko2024vera}.
LoRAForge instead starts from trained adapters.
PARA studies post-hoc rank selection~\citep{kumaravelu2026para}; FLoRIST,
SpecTraL, and FraQ provide factor-space recompression
\citep{ramesh2026florist,ramesh2026spectral,li2026fraq}.

\paragraph{Compression of adapter collections.}
Compress-then-Serve jointly compresses adapter collections
\citep{brueelgabrielsson2024compressthenserve}; Compress-then-Merge produces
one merged adapter~\citep{compressthenmerge2026}.
LoRAForge preserves independent adapters under a shared budget and carries
retained ranks into execution. FlashTSQR builds on FraQ and
communication-avoiding QR~\citep{li2026fraq,demmel2012tsqr}, contributing
batched GPU factor construction.

\paragraph{Serving and hardware-aware execution.}
Orca introduces iteration-level batching~\citep{yu2022orca}; vLLM pages
KV caches~\citep{kwon2023vllm}; SGLang reuses prefixes~\citep{zheng2024sglang}.
Sarathi-Serve chunks prefills~\citep{agrawal2024sarathi}, while DistServe
separates prefill and decode~\citep{zhong2024distserve}.
Punica, S-LoRA, dLoRA, and Toppings optimize adapter kernels, memory, and
scheduling~\citep{chen2024punica,sheng2024slora,wu2024dlora,li2025toppings}.
FlashInfer specializes attention kernels~\citep{ye2025flashinfer}, and
HAQ explores hardware-aware compression~\citep{wang2019haq}.
LoRAForge specializes shrink and expand for heterogeneous compressed ranks.

\section{Conclusion}
\label{sec:conclusion}

LoRAForge connects batched factor construction, retained-rank representation,
and specialized execution in one system. Uniform retention and spectral
allocation share this machinery, and both benefit from the same runtime in
matched-work controls. Their quality--resource tradeoffs guide policy choice. Frozen spectral deployments
achieve $1.59\times$/$1.92\times$ token throughput on MI250X/GH200;
these gains include execution and residency effects.

\section*{AI Usage Disclosure}
We used OpenAI Codex and ChatGPT to assist with manuscript editing,
literature searches, experimental and plotting scripts, visualization of
experimental results, and submission preparation. These tools helped revise
wording, identify potentially relevant literature, develop and revise scripts,
and prepare figures from experimental outputs. Reported measurements come
from experimental runs, rather than AI-generated data. The authors are
responsible for the paper's claims, references, code, and results.
\clearpage
\appendix

\raggedbottom
\captionsetup{font=footnotesize,skip=4pt}
\setlength{\textfloatsep}{12pt plus 2pt minus 2pt}
\setlength{\dbltextfloatsep}{12pt plus 2pt minus 2pt}
\setlength{\floatsep}{8pt plus 2pt minus 2pt}
\setlength{\dblfloatsep}{8pt plus 2pt minus 2pt}
\setlength{\intextsep}{10pt plus 2pt minus 2pt}
\renewcommand{\topfraction}{.92}
\renewcommand{\dbltopfraction}{.92}
\renewcommand{\textfraction}{.08}
\renewcommand{\floatpagefraction}{.75}
\renewcommand{\dblfloatpagefraction}{.75}
\setcounter{topnumber}{3}
\setcounter{totalnumber}{4}

\section{Implementation Details}
\label{app:implementation}

\subsection{Factor Construction and Manifest Details}
\label{app:factor-construction}

\paragraph{One-sided factorization assumptions.}
For $\Delta\mat W=s\mat B\mat A$, factor construction assumes
$r\le\min(d_{\mathrm{in}},d_{\mathrm{out}})$. It factors the side with
smaller ambient dimension, so $\mat X\mat Y$ is $\mat B\mat A$ or its
transpose. Orthogonality of $\mat Q$ means that
$\mat H=\mat R\mat Y$ has the same nonzero singular values as
$\mat X\mat Y$. Its core EVD therefore orders the directions used by
Equation~\eqref{eq:flash-reconstruction}. The allocator uses the effective
spectrum $|s|\sigma_j$, including rank-stabilized scaling when configured;
exports preserve the original effective update scale.

\paragraph{Leaf and merge implementation.}
For $M$ matrices $\mat X_m\in\mathbb R^{d_s\times r}$, where
$d_s=\min(d_{\mathrm{in}},d_{\mathrm{out}})$, a tile height $b\ge r$
gives $P=\lceil d_s/b\rceil$ tiles per matrix. A leaf kernel launches
$MP$ GPU blocks, zero-pads a partial final leaf, and carries an unpaired
node to the next tree level. At zero-based column $j$ of a triangular
merge, only $j+2$ entries participate in the Householder vector. The two
upper triangles occupy $r(r+1)$ packed floats, and intermediate
$\mat R$ factors remain packed between levels.
When the default tiles exceed the device's shared-memory capacity, dispatch
selects a separate Householder TSQR kernel with global-memory tiles. This
wider-rank path uses dense merges rather than the default packed merge.

\paragraph{Mixed precision and eigensolver dispatch.}
TSQR and implicit-$\mat Q$ application use FP32. The implementation
casts $\mat R$ and $\mat Y$ to FP64 before forming both
$\mat H=\mat R\mat Y$ and $\mat G=\mat H\mat H^\top$,
and performs the EVD in FP64. This reduces additional rounding error
in the Gram computation, although it does not remove the conditioning
penalty of squaring the singular spectrum or the FP32 QR error.
Eigenvalues are clamped to nonnegative values before taking square roots,
and near-zero directions receive a guarded inverse scaling.
On CUDA, the core EVD uses the library symmetric eigensolver. On ROCm,
the current dispatcher uses a custom batched FP64 cyclic Jacobi kernel
when its workspaces fit in shared memory, and a CPU FP64 EVD otherwise.
The Jacobi path assigns one workgroup to each core matrix and rotates
disjoint index pairs in parallel within each sweep.

\paragraph{Reconstruction entry points.}
The refactor entry point applies implicit $\mat Q$ to all $r$ columns of
$\mat V\mat\Sigma^{1/2}$ and returns the full ordered factors and spectrum.
The lower-level apply operator accepts a narrower matrix, but the current
entry point does not defer reconstruction until after rank selection.
Full refactoring is batched across modules; the single-module truncation
wrapper invokes the same entry point with $M=1$ and zero-pads its result.

\paragraph{Manifest provenance and numerical validation.}
The recorded compressed exports use the custom \texttt{tsqr\_full} or
\texttt{tsqr\_wide} construction path.
Manifests record original/cache hashes, kernel provenance, module ranks,
compact cost, and physical tensor sizes. These records bind each export to
its original weights and construction path. Validation scripts compare
reconstruction actions against original factors, spectra against dense FP64
references, and QR reconstruction and orthogonality. The recompression experiments
report numerical error and timing under their respective precision and backend
settings (Section~\ref{sec:conversion-latency} and
Appendix~\ref{app:mi250x-conversion}).

\subsection{Retained-Rank Execution Implementation and Validation}
\label{app:execution-implementation}

\paragraph{Native-kernel provenance.}
Our tile-size observations refer to inspected dense-LoRA source snapshots:
vLLM revision \texttt{974927f} and SGLang revision \texttt{4372829}.
The vLLM defaults use a rank-axis output tile of 16 for shrink and a
reduction tile of 32 for expand; SGLang CSGMV uses 16 for both.
Masked accesses and rank-bounded loops preserve correctness, but partially
filled tiles can still perform unnecessary work. These are properties of
the inspected defaults; both implementations support tuned configurations.

\paragraph{Platform mappings and precision.}
The custom shrink kernel and RowExpand support HIP on AMD and CUDA on NVIDIA.
They use BF16 factors and activations with FP32 accumulation, targeting retained
ranks 1--64. Shrink maps one token and one retained row of $A$ to a 64-thread
wavefront on AMD or a 32-thread warp on NVIDIA. Threads load eight contiguous
BF16 elements at a time, accumulate products in FP32, and reduce through the
platform's shuffle operations. Rank metadata terminates groups beyond the
effective rank before operand reads; packed slice offsets identify the
appropriate direction in fused projections.

\paragraph{RowExpand buckets and work scheduling.}
For retained rank $k_m$, RowExpand uses
$\rho_m=2^{\lceil\log_2 k_m\rceil}\in\{1,2,4,8,16,32,64\}$,
zero-padding unused entries. These compile-time widths specialize the local
reduction and weight loads; larger buckets use vectorized loads. Each work
item covers up to 16 token rows from one adapter segment and a block of output
channels. The kernel applies LoRA scaling and accumulates into the base output,
supporting up to three fused projection slices. A fixed grid of persistent
blocks strides over live (segment, output-column-block) work items using a
device-side segment count, without a separate launch per rank bucket.

\paragraph{Compact access and device metadata.}
The native pool remains available for loader and CSGMV compatibility.
Beside it, a persistent compact $B$ shadow stores each fused projection and
adapter in a $[d_{\mathrm{out}},\rho_m]$ region with a fixed device offset.
Per-slot device metadata stores rank, bucket, and offset. Slot loading
refreshes the metadata and shadow without changing their addresses, preserving
captured CUDA/HIP graphs under changing adapter maps. This shadow is additional
resident storage; it does not compact or replace the native pool.
In the backend screening configuration, the compact $B$ shadow occupies
594,271,232 bytes, compared with about 1.8\,GB for the earlier fixed-16
shadow. The native pool remains resident.

Fused QKV and gate/up operations use the largest retained slice rank as their
effective rank; shorter slices remain zero-padded within the operation.
The corresponding $A$ slices are repacked from checkpoint stride to effective
stride before use. The rank-aware CSGMV shrink exits column blocks beyond
that effective rank before the reduction loop, avoiding tensor-core work on
zero-masked columns. The unmodified CSGMV baseline does not include this change.

\paragraph{Frozen serving policy.}
In the evaluated MI250X configuration, decode uses RowExpand through rank 64,
while a frozen table selects shrink by projection type and captured
total-token/batch shape. Prefill uses rank-aware CSGMV shrink and RowExpand
only through rank 16. The table was selected using real factor shapes and
synthetic activations, then checked end to end; it does not inspect task
labels. Device-side rank masks route unsupported ranks to native kernels,
including the uncompressed control's rank-128 adapter. Rank-31--64
HypergraphFormer modules use RowExpand during decode and native prefill
expand. The policy is transferred unchanged from Qwen to Llama; it is not
retuned on the rank/batch sweep. Appendix~\ref{app:factorial} records the
historical input exposure during policy selection.

\paragraph{Graph replay and numerical checks.}
The timed path contains no GPU-to-CPU rank reads or rank-count atomics.
Integration checks cover nonzero base-output accumulation, mixed adapters
including the unsupported-rank fallback, and graph replay after changing
the adapter map. Each Qwen50 serving configuration additionally checks
single-request eager/graph consistency for all 50 adapters at synthetic
lengths 131, 255, and 257. BF16 factors and outputs with FP32 accumulation
can produce small numerical differences from CSGMV; these correctness checks
do not establish task-quality equivalence across backends.
Appendix Sections~\ref{app:rank-kernels} and~\ref{app:expand-ablation} specify
operator acceptance thresholds, while Appendix~\ref{app:factorial} details
the serving validation protocol. Section~\ref{sec:quality-budgets} reports
the full task-quality evaluation.

\section{Supplementary Experiments}
\label{app:experiments}

This appendix collects the protocols, additional measurements, and diagnostics
supporting the main evaluation. It includes recompression, serving controls,
operator sweeps, and per-adapter quality regressions. Factor-construction and
runtime implementation details are collected separately in
Appendix~\ref{app:implementation}.

\paragraph{Policy and checkpoint scope.}
Uniform and Spectral in crossed controls denote rank-allocation policies;
Native and LoRAForge denote execution backends. Uniform and Spectral with
the shared runtime correspond to \lfuniform{} and \lfspectral{}.
At $1/12$, Table~\ref{tab:quality-fleets}, the Qwen control in
Figure~\ref{fig:allocation-execution}, and the AMD and GH200 deployment
curves use the same frozen rank-floored spectral exports for each model.
The rank floor is $\ell_m=\max(1,\lceil u_m/2\rceil)$; the corresponding
Qwen/Llama quality scores are 69.41/69.93. The w/o-floor results are a
quality-only ablation, not the serving configuration.
Uniform's direct runtime control covers Qwen at one budget and two loads.

\begin{table}[!htbp]
\centering
\caption{\textbf{Shared Spectral checkpoint configuration.} All rows use
budget $1/12$ and rank floor $\ell_m=\max(1,\lceil u_m/2\rceil)$.
Each model's compressed weights are shared across its quality and serving runs.}
\label{tab:checkpoint-scope}
\footnotesize
\setlength{\tabcolsep}{4pt}
\begin{tabular}{lll}
\toprule
Experiment & Cohort & Spectral exports \\
\midrule
Table~\ref{tab:quality-fleets}, $1/12$ & Qwen50 / Llama35 & Shared per model \\
Figure~\ref{fig:allocation-execution} & Qwen50 & Same as quality \\
AMD deployment & Qwen50 / Llama35 & Same as quality \\
GH200 deployment & Qwen50 / Llama35 & Same as AMD \\
\bottomrule
\end{tabular}
\end{table}

\subsection{vLLM Mixed-Fleet Serving}
\label{app:vllm-serving}
\begin{figure*}[!t]
\centering
\includegraphics[width=\linewidth]{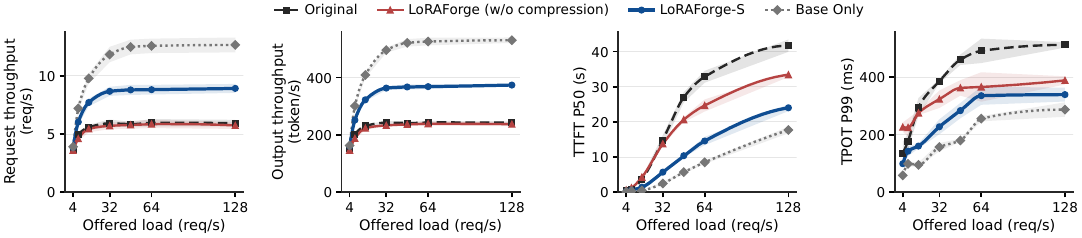}
\caption{\textbf{Four-way mixed-fleet serving comparison in vLLM on MI250X.}
All configurations share a four-decode-step scheduling policy.
Markers show five-seed means and bands sample SD over 1,700 requests per seed;
smooth lines guide the eye. LoRA configurations use natural stopping;
Base Only matches \lfspectral{}'s per-request output lengths.}
\label{fig:mixed85-vllm}
\end{figure*}

This experiment tests whether the end-to-end benefit in
Section~\ref{sec:exp-serving} extends to another serving engine.
Figure~\ref{fig:mixed85-vllm} repeats the four configurations of
Figure~\ref{fig:mixed-serving} entirely within vLLM: Original uses native LoRA
kernels; LoRAForge (w/o compression) applies the optimized execution path to
original adapters; \lfspectral{} combines that path with the frozen fleet-compressed
factors; and Base Only disables LoRA.

\paragraph{Matched configuration.}
Two vLLM 0.26.0 engines co-locate Qwen50 and Llama35 on one MI250X GCD.
We reuse the AMD experiment's adapters and frozen inputs, with BF16, tensor
parallelism one, 16,384 KV tokens per model, and 14/10\,GiB Qwen/Llama
adapter-storage ceilings including compact $B$ shadows.
The retained-rank interface and RowExpand are adapted to vLLM's separate
projection slices. All four configurations use the same asynchronous scheduler
extension: after a prefill-containing step, it defers new admissions and partial
prefills for four nonempty decode steps when decode work is available.
Prefill-release steps retain vLLM's mixed batching.
Five seeds each issue 1,700 requests at seven offered rates from 4 to 128
requests/s, giving 140 measured points. Every point follows an 85-request
warm-up and a full 1,700-request conditioning pass. Base Only replays the
corresponding vLLM \lfspectral{} output lengths with exact caps and EOS ignored;
all configurations use fresh HTTP connections without per-request retries.

\paragraph{Throughput and latency.}
At 32, 48, 64, and 128 requests/s, \lfspectral{} improves token throughput over
Original by $1.50\times$, $1.52\times$, $1.51\times$, and $1.54\times$,
respectively. At 128 requests/s, throughput increases from 242.8 to
374.0 tokens/s, TTFT P50 falls from 41.7 to 24.0\,s, and TPOT P99 falls from
513.1 to 339.8\,ms (a 33.8\% reduction).
Thus the compressed serving representation improves both throughput and
latency under this second engine's execution and scheduling stack.

\paragraph{Why compression and execution remain coupled.}
Applying the optimized path to uncompressed adapters yields only
$0.96$--$0.98\times$ Original's token throughput at 32--128 requests/s,
although it reduces high-load TPOT P99. At 128 requests/s, this configuration
reaches 237.4 tokens/s and 389.3\,ms, versus \lfspectral{}'s 374.0 tokens/s and
339.8\,ms. The results support transferring the complete compressed
representation and execution path, rather than attributing the gain to a
kernel replacement alone. They measure the combined effects of compression,
residency, execution, and natural output lengths; they do not isolate a
RowExpand-only speedup.

\paragraph{Scope of comparison.}
These are within-vLLM comparisons. The scheduler extension approximates the
SGLang decode cadence while retaining vLLM's own mixed batching, and per-point
conditioning differs from the SGLang sweep; the two figures therefore do not
rank the engines against each other. Base Only reaches 530.5 tokens/s at
128 requests/s, but its exact caps and ignored EOS also affect admission
(Appendix~\ref{app:mixed-tpot}).

\subsection{Matched Full-Fleet Recompression on GH200}
\label{app:gh200-conversion-protocol}

This experiment supports Figure~\ref{fig:conversion-gh200} and
Section~\ref{sec:conversion-latency}. It isolates the effect of replacing
library QR inside the same batched factor-construction pipeline.

We use one GH200 (95\,GiB visible memory), PyTorch 2.11.0/CUDA 13.0,
and seven CPU threads per job. Original scaled inputs match frozen SHA256
manifests: 12,492 Qwen50 and 7,840 Llama35 matrices, ranks 16--128.
These are local algorithmic references, not official author implementations.

\paragraph{Matched execution and precision.}
FLoRIST performs two factor SVDs and one core SVD on each two-dimensional
A/B pair, using FP64 and casting final factors to FP32. SpecTraL uses two
FP32 QRs, GPU FP64 core SVD, and FP32 reconstruction per pair. Neither
reference uses batched factorizations here.
Both FraQ arms share B-side FP32 QR, FP64 Gram/reconstruction, CUDA FP64
\texttt{cusolverDnXsyevBatched}, full-energy rank masks, and FP32 outputs.
The eigensolver uses reusable workspaces and device-side convergence checks;
there is no host EVD. Identical shape batches capped at 24 give 524/334
calls for Qwen/Llama. Only QR changes: \texttt{torch.linalg.qr} versus
our \texttt{qr\_factor}, recording \texttt{tsqr\_full} for every shape.
FlashTSQR uses its original leaf application and materializes Q through
\texttt{applyQ(identity)}. Both arms then perform the same FP64 Q--eigenvector
multiply. This measures QR replacement within stabilized FraQ, not fused
production applyQ or unmodified upstream FP32 arithmetic.

\paragraph{Timing, acceptance, and interpretation.}
Inputs and full-rank output buffers reside on the GPU before two warmups
and five synchronized passes. Timing includes allocations, eigensolver calls,
reconstruction and output copies; initial IO/upload, JIT and validation are
excluded. Every matrix passes action error $<3\times10^{-5}$ and spectrum
error $<10^{-4}$ in all eight results. Each method/fleet runs in an independent
one-GPU job; repeat ranges are descriptive, not confidence intervals.
The compressed exports used in the quality and serving experiments are
reconstructed from hash-verified full-rank FlashTSQR caches, which pass the
recorded action-error threshold of $3\times10^{-5}$ before prefix export.
These numerical checks validate factor construction, not downstream task quality.

The matched control materializes full ordered factors before rank selection.
It does not time allocation, checkpoint export, or serving. Shared batched
EVD benefits both FraQ arms; the reported ratios measure factor-construction
time, not isolated QR-kernel latency. Earlier GH200 EVD results and
failed FP32 controls remain archived. Data, source hashes and scripts:
\path{evidence/20261029/conversion-gh200-batched-evd}.

\subsection{Full Rank and Kernel Crossover Measurements}
\label{app:rank-kernels}

\paragraph{Shrink-only aggregation.}
Figure~\ref{fig:shrink-ablation} reuses the archived per-stage timings,
without new GPU runs. For each fleet it includes ranks
$\{1,2,4,8,12,16\}$ and all four projection shapes. Decode uses token counts
$\{1,8,32,128,256\}$ (120 cases per bar); prefill uses
$\{128,512,2048,8192\}$ (96 cases per bar). Each ratio divides the native
shrink median by the tested shrink median for identical operands, rank,
pool capacity, and routing; bars take their equal-weight geometric mean.
The underlying medians use five timing blocks. No confidence interval or
workload-frequency weighting is implied. The fixed-128 control measures
effective-rank pruning under padding; it is not the measured pool layout of
every serving deployment. Source hashes and all 624 plotted case ratios are
recorded in \path{evidence/20261030/shrink-ablation}.

Rank 33 is omitted from plotted comparisons pending alignment analysis;
its measurements remain in the archived 2,016-case sweep.

We freeze the same CSGMV and HIP implementations and dispatch table used by
the serving experiments. The baseline invokes unmodified SGLang CSGMV;
the rank-aware shrink is part of the optimized path. Controlled BF16 weights
and activations use the actual Qwen/Llama QKV, O, Gate/Up, and Down dimensions.
All ranks take prefixes of fixed rank-128 master factors. The operands,
adapter mapping, and scaling match across paths and pool layouts. Up to eight
adapters are active, depending on batch size. Tight pools have capacity equal
to retained rank; the capacity-128 control retains native padding while the
HIP path uses its additional compact B storage.

Each case validates all live outputs against native CSGMV and sampled rows
against FP32 matrix products with a BF16 intermediate. Acceptance requires
relative L2 error below 1\% and maximum absolute error below 2.5\% of
$\max(0.1,\|y_{\mathrm{ref}}\|_\infty)$. All 2,016 cases pass.
Five interleaved, randomly ordered timing blocks use GPU events and graph
replay; compilation, preparation, and validation are excluded. Warm-cache
measurements use repeated replay. The cache-flushed figures below fill a
128\,MiB device buffer outside the timed interval before each replay; this
does not claim to reproduce a full model's cache behavior. Per-shape timings
are retained, while figures use equal-weight geometric means across four
projection types. Forced HIP is unsupported above rank 64 and is marked N/A;
the automatic path includes its native fallback and associated overhead.

\begin{figure}[!htbp]
\centering
\includegraphics[width=\linewidth]{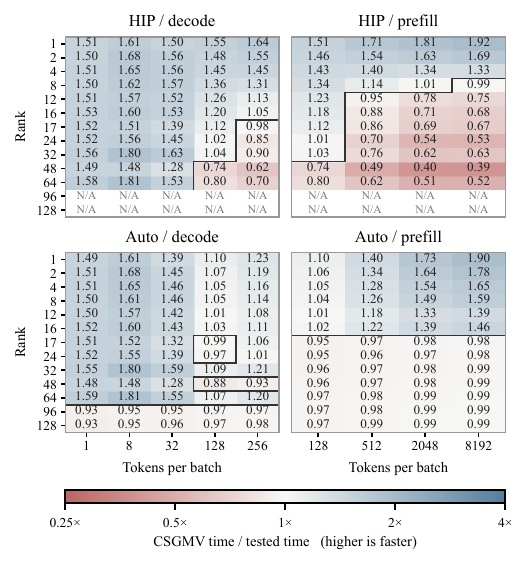}
\caption{\textbf{Rank and batch crossovers on Llama35.}
Tight rank capacity, with forced HIP (top) and the frozen automatic policy
(bottom), complementing the Qwen results in Figure~\ref{fig:rank-kernel}.
Values are geometric-mean native/tested shrink-plus-expand latency ratios
across four projection shapes with cache flushing. Rank 33 is omitted;
forced HIP is unsupported above rank 64. Black boundaries mark $1\times$
crossovers between measured cells.}
\label{fig:rank-llama-tight}
\end{figure}

\begin{figure}[!htbp]
\centering
\includegraphics[width=\linewidth]{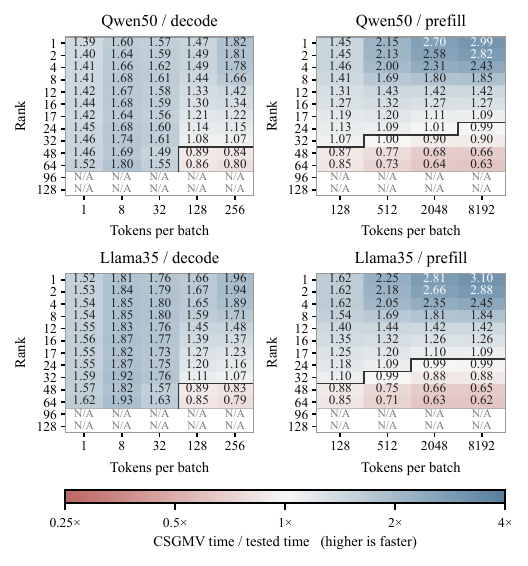}
\caption{\textbf{Forced HIP with a native pool capacity of 128.}
The mathematical retained rank is unchanged; this control exposes the
combined effects of native padding, kernel execution, and compact B storage.}
\label{fig:rank-full-hip-padded}
\end{figure}
\begin{figure}[!htbp]
\centering
\includegraphics[width=\linewidth]{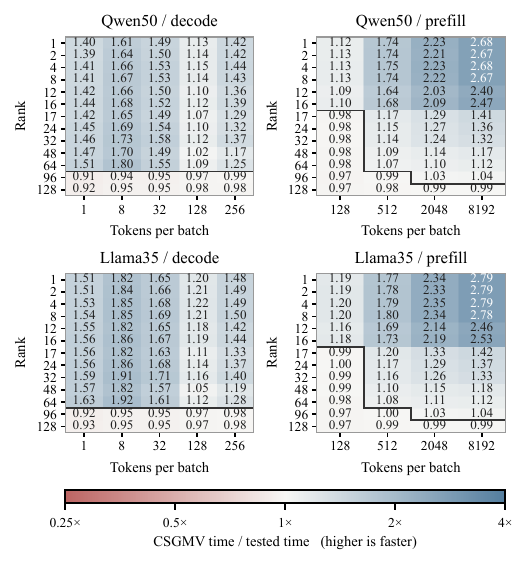}
\caption{\textbf{Frozen automatic policy with a native pool capacity of 128.}
Both fleets follow the same previously frozen policy. These operator ratios
do not include model execution, adapter loading, or CPU--GPU paging.}
\label{fig:rank-full-auto-padded}
\end{figure}
\subsection{Allocation--Execution Control Protocol}
\label{app:factorial}

Figure~\ref{fig:allocation-execution} presents the throughput comparison.
Table~\ref{tab:factorial-latency} adds latency and allocated-memory measurements;
the protocol below specifies the matched workload and execution controls.

\begin{table}[ht]
\centering\fontsize{7}{8.5}\selectfont
\setlength{\tabcolsep}{2.5pt}
\captionsetup{font=footnotesize}
\caption{\textbf{Latency and allocated-memory footprint in the six-cell control.}
Latency columns give the arithmetic mean of each run's quantile across three
input seeds, not a pooled quantile. Column labels 16/32 are offered requests/s.
Peak is the mean PyTorch allocated-memory high-water mark over each engine's
entire lifetime, including loading, validation, and graph capture; each engine
serves both rates. It is neither timed-phase-only memory nor device-wide usage.}
\label{tab:factorial-latency}
\begin{tabular}{llrrrrr}
\toprule
Allocation & Runtime & \multicolumn{2}{c}{TTFT P50 (s)} & \multicolumn{2}{c}{TPOT P99 (ms)} & Peak (GiB) \\
 & & 16 & 32 & 16 & 32 & allocated \\
\midrule
Original & Native & 1.74 & 14.77 & 245.1 & 286.4 & 45.88 \\
Original & LoRAForge & 1.46 & 13.85 & 224.0 & 274.0 & 51.62 \\
Uniform & Native & 0.73 & 8.95 & 170.7 & 227.0 & 23.88 \\
Uniform & LoRAForge & 0.41 & 6.67 & 147.4 & 203.7 & 24.60 \\
Spectral & Native & 0.96 & 10.42 & 192.5 & 242.4 & 33.30 \\
Spectral & LoRAForge & 0.57 & 8.33 & 154.5 & 222.5 & 33.88 \\
\bottomrule
\end{tabular}

\end{table}

\paragraph{Workload and stopping contract.}
Each input seed defines 1,000 requests covering the 50 Qwen adapters. Every
allocation/path pair replays the same input identities, token sequences,
adapter order, seeded Poisson arrivals, and completion-length vector. Lengths
come from the previously recorded, same-seed Spectral natural-EOS run at offered
rate 48, frozen before this control and reused at both tested rates.
For all six cells, \texttt{min\_tokens} equals \texttt{max\_tokens},
\texttt{ignore\_eos=True}, and \texttt{skip\_special\_tokens=False}.
We verify all 36,000 completion counts and six-way request, arrival, and
workload hashes. This controls generated counts and stopping semantics,
not generated content, task correctness, or the realized schedule.
Throughput includes draining the finite request stream; it is not inferred
from the offered rate or an isolated steady-state decode window.

\paragraph{Engine and allocation controls.}
Each engine uses one MI250X GCD, tensor parallelism one, BF16 serving,
context length 16,384, maximum 256 running requests, prefill chunk 8,192,
static memory fraction 0.8, and prefill/decode interval four. All adapters
are pinned, with 51 slots including the null adapter. Radix caching is off;
graph and normalization settings are shared. The 86,000-token KV budget
fits Original with optimized execution, the most memory-heavy arm. Historical
automatic capacities differed by configuration, so these timings must not be
compared directly to the natural-EOS curves as if capacity were unchanged.

All checkpoints are reused byte-for-byte after source-hash verification.
Original has 5,946,605,568 parameters; Uniform and Spectral have 457,869,312
and 457,866,752 compact parameters. Checkpoint padding and native pool
capacity are recorded separately. Original/Uniform/Spectral use native pool
rank capacities 128/16/64, respectively, identical within each native/optimized
pair. Native retains its loader-compatible padding. Optimized execution adds
effective module-rank handling, compact B access, and the frozen dispatch;
the native pool remains resident. Thus the execution comparison deliberately
measures the full production path, not HIP arithmetic alone.
Table~\ref{tab:factorial-latency} shows that optimized Spectral's allocated peak
is 33.88\,GiB versus Uniform's 24.60\,GiB despite their matched compact budgets.
The spectral budget is not a physical-memory or latency constraint.

Within Spectral, the retained HypergraphFormer adapter receives ranks 31--64
across modules and 130,387,456 compact parameters, about 28.5\% of the
total budget. This concentration follows weight spectra, not task
calibration or request-length weighting.

\paragraph{Validation, timing, and pairing.}
Eighteen independent engines each pass mixed-adapter integration checks and
150 single-request eager/graph comparisons (50 adapters at lengths 131,
255, and 257) before timing. A short warmup and a complete 1,000-request
conditioning pass precede both measured rates. Seed 102 reverses rate order
for all cells. Native/optimized pairs share a node on distinct GCDs; eight
engines run on each of two full nodes and the remaining pair on a third.
Every measured point completes with no failed requests. Runtime-source
hashes, closed-audit acknowledgments, route audits, and exact output counts
are checked. Three input seeds provide paired workload observations, not
three independent timing repetitions of one workload. We report all seeds;
neither seed ranges nor standard deviations are confidence intervals.

\paragraph{Relationship to earlier execution controls.}
The frozen policy was selected using input seed 101; seeds 102/103 provide
additional input samples from the historically exposed pools. It is unchanged
in all six cells. The uncompressed control does not exhaust high-rank tuning.
These fixed-work ratios cannot be multiplied with natural-EOS deployment ratios.

Policy screening used eight variants at seed 101 against the previous
packed-hybrid compressed backend. Its later confirmation (seeds 102/103,
two repeats, rates 16/32) gave a geometric-mean ratio of $1.198\times$ over
that older compressed reference. An earlier uncompressed native comparison
gave $0.965\times$. Those controls used different references and automatic
KV capacities; their ratios are not the cells of the new factorial control.
Figure~\ref{fig:allocation-execution} replaces them as the primary matched
comparison without changing the frozen policy. Natural-EOS serving curves
measure complete deployment behavior. This control uses the same rank-floored $1/12$ spectral exports as
Table~\ref{tab:quality-fleets} (69.41 quality points on Qwen).

\paragraph{Reproduction.}
The 36 compressed raw result files, manifests, validation records, source
hashes, and submission receipts are archived under
\path{evidence/20261029/qwen50-factorial}. The source snapshots record the
client contract and execution policy. Running
\path{scripts/plot_qwen50_factorial.py} audits the raw results against their
SHA256 hashes, recomputes throughput, checks six-way workload matching, and
regenerates the figure, latency table, and paired-ratio summaries.

\subsection{Fixed-Work Controls at 64 Requests/s}
\label{app:fixedwork64}

\begin{figure}[!htbp]
\centering
\includegraphics[width=\linewidth]{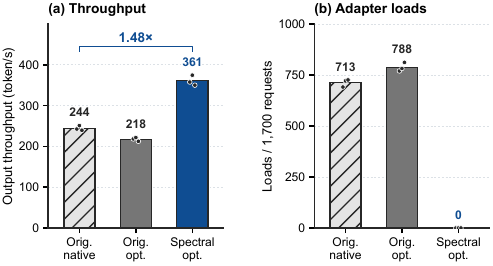}
\caption{\textbf{Mixed-fleet gains under matched output work.}
MI250X at 64 requests/s, three paired seeds with identical inputs, arrivals,
output lengths, and stopping rules. Bars show means and dots seeds.
(a) Token throughput; the bracket gives the geometric mean of paired
\lfspectral{}/Original-native ratios. Here opt. denotes the LoRAForge runtime.
(b) Adapter loads during the same runs.
KV and adapter-storage budgets are unchanged.}
\label{fig:fixedwork64-mixed}
\end{figure}

\begin{table}[ht]
\centering\fontsize{7}{8.5}\selectfont
\setlength{\tabcolsep}{2.8pt}
\captionsetup{font=footnotesize}
\caption{\textbf{Fixed-work throughput, latency, and paging.}
Identical complete seed sets are used for all arms within each scenario.
Latency entries average per-run quantiles, not pooled request quantiles.
Loads and H2D source bytes count only the timed request window; source bytes
are not a measurement of physical PCIe traffic or transfer latency.
Llama rows are provisional; Mixed85 rows include all three planned seeds.}
\label{tab:fixedwork64}
\begin{tabular}{@{}lrrrrr@{}}
\toprule
Path & tok/s & TTFT & TPOT & Loads & H2D \\
 & & P50 (s) & P99 (ms) & & GiB \\
\midrule
\multicolumn{6}{@{}l}{\textit{Llama35: 2 paired seeds, provisional}} \\
Orig./native & 421.5 & 7.96 & 466.8 & -- & -- \\
Orig./LoRAForge & 439.9 & 6.78 & 429.6 & -- & -- \\
Spectral/LoRAForge & 559.6 & 5.05 & 369.7 & -- & -- \\
\midrule
\multicolumn{6}{@{}l}{\textit{Mixed85: 3 paired seeds, complete}} \\
Orig./native & 244.4 & 41.95 & 390.1 & 713 & 336.8 \\
Orig./LoRAForge & 217.5 & 39.07 & 313.4 & 788 & 363.3 \\
Spectral/LoRAForge & 360.9 & 12.81 & 254.6 & 0 & 0.0 \\
\bottomrule
\end{tabular}

\end{table}

The planned control has 18 points: Llama35 or Mixed85, three deployed arms,
and seeds 101--103, all at offered rate 64. Llama sends 700 requests per point;
Mixed85 sends 1,700, preserving the 50:35 model mix. This archive contains
16 completed points with no failed requests. Table~\ref{tab:fixedwork64}
uses 15 points: all nine Mixed85 measurements and all three Llama arms for
seeds 101/102. Figure~\ref{fig:fixedwork64-mixed} shows the nine Mixed85 points. The completed, unpaired Spectral
seed 103 is retained in the archive but excluded from those Llama aggregates;
its two Original counterparts were queued. No missing value is imputed.

\paragraph{Common work and preserved execution policy.}
Within each scenario/seed, all arms reuse the request identities, tokenized
inputs, order, arrivals, and frozen completion lengths of the historical
Spectral run at rate 64. Minimum and maximum lengths are equal, EOS is ignored,
special tokens are retained, and custom stop-token IDs are omitted for all
arms. Exact per-request counts and request/arrival hashes are checked.
The checkpoint bytes, rank allocation, production HIP--CSGMV policy, graph
settings, BF16 arithmetic, and prefill/decode interval four are unchanged.
Each engine warms up and runs a full conditioning pass before its one formal
measurement. Numerical and eager/graph checks precede timing; Mixed85 also
checks eviction/reload roundtrips. These are execution checks, not task scores.

\paragraph{Memory and residency controls.}
Single-model Llama pins all 35 adapters with 36 slots and fixes KV capacity
at 76,000 tokens, below the historical Original/optimized capacity of 78,288.
Historical Original/native and Spectral capacities were larger. This control
therefore changes both stopping semantics and KV capacity relative to
Figure~\ref{fig:serving-llama}; it is not an output-length-only intervention.

Mixed85 retains 16,384 KV tokens per model and the 14/10\,GiB LoRA ceilings.
For native Original, optimized Original, and optimized Spectral, Qwen has
28/16/51 native-pool slots and Llama has 15/9/36. These counts follow each
representation's memory cost including B shadows, rather than equalizing
resident-adapter counts and removing the compression benefit. All 85
adapters remain accessible through CPU-to-GPU paging. Native-pool and shadow
bytes are checked against the ceilings. Counter snapshots taken outside the
timed window give adapter-load and copied-source-byte deltas. Host load/enqueue
times are also archived, but are not GPU-event transfer times. No independent
paging-disable control isolates transfer savings from execution changes.

\paragraph{Additional results and completion status.}
Llama's two complete paired seeds yield 421.5/439.9/559.6 tok/s for native
Original, optimized Original, and optimized Spectral: a paired Spectral/native
Original gain of $1.33\times$. The unpaired Spectral seed 103 is excluded while
its two Original counterparts remain queued at this archive snapshot.
For Mixed85, native Original copies 336.8\,GiB of source tensors on average.
Optimized Original averages 788 adapter loads and 217.5 tok/s; its B shadow
leaves fewer native-pool slots under the same storage ceilings. Source-byte
counters do not measure physical link traffic or time spent transferring.
The fixed-work $1.48\times$ and natural-EOS $1.59\times$ gains are system ratios;
their difference is not an isolated estimate of output-length effects.

\paragraph{Reproduction and scope.}
\path{evidence/20261029/fixedwork64} retains all 16 completed raw results,
content-addressed inputs, validation and paging snapshots, source provenance,
and pending/unpaired status. \path{scripts/plot_fixedwork64.py} verifies hashes,
request identities, output counts, arrival matching, throughput, quantiles,
storage ceilings, and paging deltas before plotting the declared paired seed
sets. The Mixed85-only view is rendered by
\path{scripts/plot_compact_execution_controls.py}.
The historical five-seed natural-EOS curves remain unchanged. Neither
their ratios nor the new control ratios establish a latency SLO, identical
generated content, or optimized-backend task-quality equivalence.

\subsection{RowExpand Ablation Protocol and Resource Usage}
\label{app:expand-ablation}

\begin{figure}[!htbp]
\centering
\includegraphics[width=\linewidth]{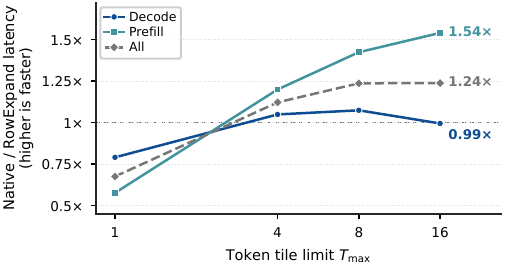}
\caption{\textbf{RowExpand versus native CSGMV.}
Geometric means of native/staged-RowExpand latency ratios over 128 decode
and 128 prefill cases; higher is better. $T_{\max}$ caps same-adapter token
rows per work item. Bands span five timing-block aggregates of cache-flushed
measurements. Token-reuse and staging ablations appear in
Figure~\ref{fig:expand-ablation}.}
\label{fig:expand-native}
\end{figure}

The 256 conditions span two model geometries, four projection types,
ranks 4/16/32/64, decode batches 32/128 or prefill batches 512/2048,
and one or eight adapters. Each compares tile limits $1,4,8,16$ with and
without staging, plus production and native references. The production
reference agrees with staged $T_{\max}=16$ within 0.1\% in aggregate latency.

We derive the experimental kernels from the frozen serving
\texttt{fast\_decode.cu}, preserving compact B layout, vectorized B loads,
reduction order, scaling, BF16 output rounding, and the persistent launch of
880 blocks with 256 threads. Each of eight workers uses one MI250X GCD for
one model/projection geometry. Fixed synthetic BF16 master factors and
activations supply rank prefixes; native B capacity equals retained rank,
and all private variants use the same compact B. Decode rows are shuffled
across active adapters, while prefill rows are grouped. Tile limits alter
segment boundaries but preserve the mathematical row-to-adapter mapping.
At batch 32 with eight adapters, for example, each adapter has only four
rows, so tile limits 8 and 16 cannot expose further cross-token reuse.

The direct-Z variant retains register-resident B but reads activation
entries directly from global memory and removes shared-Z allocation/staging.
Caching may absorb repeated loads. The resource changes in
Table~\ref{tab:expand-resources} are part of this intervention, so its speed
ratio cannot be interpreted as an isolated shared-memory bandwidth effect.
The generic compiled kernel contains every rank bucket; the resource API
therefore reports one set of attributes per staging path, not per runtime
rank or tile limit. Reported occupancy is a theoretical resource limit,
not measured active-wave occupancy. No HBM traffic counters were collected.

\begin{table}[h]
\centering\small
\caption{\textbf{Compiled RowExpand resource attributes.}
HIP function and occupancy APIs, 256 threads per block. Theoretical occupancy
is 50\% for both variants (four active blocks per compute unit).}
\label{tab:expand-resources}
\setlength{\tabcolsep}{4pt}
\begin{tabular}{lrrr}
\toprule
Variant & Registers/thread & Static LDS (B) & Local (B)\\
\midrule
Shared-Z & 101 & 12,352 & 0\\
Direct-Z & 128 & 64 & 52\\
\bottomrule
\end{tabular}
\end{table}

Every timed condition validates all outputs against native CSGMV and sampled
rows against an independent FP32 matrix-product oracle with BF16 delta/update
rounding. The relative L2 threshold is 1\%, and maximum absolute error must
be below 2.5\% of $\max(0.1,\|y_{\mathrm{ref}}\|_\infty)$.
An additional 64 controls exercise ranks 1/2/4/12/16/31/32/64, partial tiles,
null slots, mixed adapters, nonunit scaling, and nonzero base outputs.
All checks pass; every private output is also bitwise equal to production.

After warmup and graph capture, five timing blocks randomize variant order.
GPU events time graph replay. Each cache-flushed sample is the median of
three replays, each preceded by a 128\,MiB flush and output reset outside
the timed interval. Warm-cache repeated-replay samples are archived separately.
Preparation, metadata construction, compilation, and validation are excluded.
For each case we divide median latencies, then take the equal-weight geometric
mean across cases; timing-block bands do not represent workload uncertainty.
Raw samples, validation results, frozen sources, and source hashes are archived
under \path{evidence/20261029/expand-ablation};
\path{scripts/plot_expand_ablation.py} recomputes the paired ratios, and
\path{scripts/plot_expand_native.py} renders the native comparison.

\subsection{Diagnosing the Co-Resident TPOT Crossover}
\label{app:mixed-tpot}

\paragraph{Parent MI250X sweep.}
The five-seed experiment in Figure~\ref{fig:mixed-serving} sweeps offered rates
$\{4,8,16,32,48,64,128\}$ requests/s, with a static memory fraction of 0.95
and prefill/decode interval four. Both engines retain 16,384 KV tokens,
with 14/10\,GiB Qwen/Llama adapter-storage ceilings including B shadows;
CPU--GPU LRU paging keeps all 85 adapters accessible. \lfspectral{} uses the
same frozen rank-floored $1/12$ spectral exports as quality evaluation.

\begin{table}[t]
\centering
\caption{\textbf{Stopping-contract controls on the same engines.}
Seed 101, 64 offered requests/s, 1,700 successful requests per row.
These are individual runs, not five-seed means. Spectral uses the optimized path.}
\label{tab:mixed-diagnostic}
\begingroup
\footnotesize
\setlength{\tabcolsep}{2.5pt}
\renewcommand{\arraystretch}{1.17}
\begin{tabular}{@{}lllrrrr@{}}
\toprule
\rowcolor{resulthead}\textbf{System} & \textbf{Stop} & \textbf{Trace} & \textbf{Output} & \textbf{TTFT} & \textbf{TPOT} & \textbf{Batch}\\
\rowcolor{resulthead} & & & tok/s & P50 (s) & P99 (ms) & peak\\
\midrule
Base & E-I & On & 417.5 & 8.64 & 212.1 & 82\\
Base & E-N & On & 467.9 & 7.96 & 221.4 & 81\\
Base & E-I & Off & 417.3 & 8.65 & 212.2 & --\\
\midrule
\lfspectral{} & N & On & 343.7 & 26.97 & 103.6 & 43\\
\rowcolor{resultours}
\lfspectral{} & E-I & On & 360.5 & 10.85 & 245.9 & 81\\
\lfspectral{} & E-N & On & 405.3 & 9.91 & 242.2 & 84\\
\lfspectral{} & N & Off & 347.6 & 26.32 & 106.3 & --\\
\bottomrule

\end{tabular}
\par\smallskip
\parbox{\linewidth}{N: task cap, natural EOS. E-I: exact cap, EOS ignored.
E-N: exact cap, normal EOS handling. Both E modes enforce the exact length
with a matching minimum. Batch peak is Llama's actual decode batch maximum
within the first 60\,s; it is unmeasured when tracing is off.}
\endgroup
\end{table}

\paragraph{Question and controls.}
The five-seed co-resident experiment shows higher aggregate throughput but
worse TPOT P99 for the length-replayed Base than for naturally stopping Spectral
at heavy offered loads. We diagnose this observation separately from the
main curves. Each arm keeps both models on one GCD and preserves the
weights, kernels, KV capacity, interval, prompts, order, and arrival schedule.
We run three Base conditions and four Spectral conditions in
Table~\ref{tab:mixed-diagnostic}, after correctness checks, a tracing smoke
test, warmup, and the same 1,700-request conditioning pass.

In N, each request has its original task cap and may finish at EOS.
In E-I and E-N, \texttt{max\_tokens} and \texttt{min\_tokens} both equal the
frozen parent Spectral output count $L_i$; E-I sets \texttt{ignore\_eos=true},
while E-N sets it to false. Both exact modes retain special tokens in the
stream. All five exact-mode runs produce identical per-request counts,
totaling 73,169 tokens. The traced natural Spectral run produces 73,323 tokens
(0.21\% more). Thus N versus E is an intervention on the length-control
contract, not an isolated change to a single flag or identical token content.

\paragraph{Why equal counts do not imply equal admission.}
In the frozen runtime, \texttt{PrefillAdder.add\_one\_req} uses the declared
output cap in its KV budget. With EOS ignored and radix reuse disabled,
it instead invokes \texttt{add\_one\_req\_ignore\_eos}, which estimates
remaining generation and memory released as requests finish.
The running-request reservation also treats fixed-length and natural-EOS
requests differently. For the parent seed-101 trace, Qwen/Llama mean declared
caps are 837.1/427.9 tokens under natural stopping, versus exact counts of
55.8/24.8 tokens under Base replay. The scheduler therefore receives
different resource estimates even when actual generated counts match.

\paragraph{Observed mechanism and intervention.}
For unchanged Spectral weights and kernels, E-I raises TPOT P99 from
103.6 to 245.9\,ms while reducing TTFT P50 from 26.97 to 10.85\,s.
The first-60-second trace shows Llama's peak decode batch increasing from
43 to 81. Its mean prefill token batch grows from 811 to 1,433, and mean
GPU-event duration for prefill rises from 168 to 307\,ms. These event
durations include execution under shared-device contention; they are not
isolated kernel timings. The trace supports the admission/batching mechanism:
more requests begin earlier, while larger prefill batches and more concurrent
generation increase the waiting experienced after the first token.

E-N leaves Spectral TPOT P99 high at 242.2\,ms; Base similarly changes from
212.1 to 221.4\,ms. Therefore changing the \texttt{ignore\_eos} flag alone,
while retaining exact caps and minima, does not remove the crossover.
Under the same E-I contract and identical output counts, Base is faster
than Spectral in both throughput (417.5 versus 360.5 tok/s) and TPOT P99
(212.1 versus 245.9\,ms). This reverses the interpretation suggested by
comparing Base E-I against Spectral N.

\paragraph{Checks and limits.}
Tracing records scheduler admissions and batches, streaming-chunk times,
and asynchronous GPU events without a per-step device synchronization.
Final untraced reruns of the original conditions reproduce Base/Spectral
TPOT P99 at 212.2/106.3\,ms, close to the initial traced 212.1/103.6\,ms.
All seven runs complete all requests with no failures, and request and
arrival hashes match across conditions. The experiment uses one seed and
one offered load, with a fixed condition order; the untraced repetitions
check drift and tracing effects but are not randomized causal replications.
It establishes sensitivity to the stopping contract on this configuration,
not its quantitative effect at every load or a pure allocation/kernel effect.
The five-seed curves remain unchanged, with the Base TPOT qualification
stated in their caption and main-text discussion.

\subsection{Full-Fleet Recompression on MI250X}
\label{app:mi250x-conversion}

\begin{figure}[!htbp]
\centering
\includegraphics[width=\linewidth]{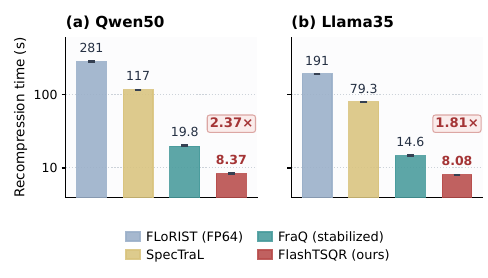}
\caption{\textbf{Full-fleet recompression time on one MI250X GCD.}
Medians and min--max whiskers over five runs; boxed ratios compare
FlashTSQR with stabilized FraQ. CPU decomposition and transfers are timed;
input I/O, upload, and JIT are excluded. Baseline definitions and numerical
checks are described in this appendix.}
\label{fig:conversion-latency}
\end{figure}

This supplementary experiment measures recompression time on AMD hardware, including
backend-dependent solver costs and CPU fallbacks. Its cross-method ratios
combine several implementation differences; the matched GH200 control in
Section~\ref{sec:conversion-latency} provides the primary QR comparison.

\paragraph{Matched workload and timing scope.}
Inputs are the original scaled factors of the served Qwen50 and Llama35
fleets: 12,492 and 7,840 matrices with original ranks 16--128.
Each pass constructs full ordered balanced factors for later prefix selection;
this workload is independent of the retained deployment budget. Adapters remain
independent, with no cross-adapter weight merging. Each method runs on one
MI250X GCD with seven CPU threads, PyTorch 2.11/ROCm 7.2, and TF32 disabled.
Inputs and FP32 output buffers are device-resident before timing. We group
matrices by exact factor shapes, splitting a group when needed to fit that
method's workspace; batch sizes therefore need not match across methods.
Two warmups precede five synchronized wall-clock measurements. Timing includes
operator allocation, factor construction, output copies, CPU fallback, and
its transfers. Initial loading/upload, shape planning, JIT, memory-fit trials, and
validation are excluded; rank allocation, prefix export, and serving are not timed.

\paragraph{Baselines and numerical acceptance.}
The FLoRIST reference computes two factor SVDs and a core SVD in FP64;
SpecTraL uses two FP32 QRs and a GPU FP64 core SVD. The FraQ reference starts
from the repository's batched one-sided implementation, retaining its B-side
library QR, grouping, and energy-based rank selection. We request all energy
with the original rank as the cap. Its native GPU FP32 EVD failed reconstruction
checks on these inputs; controlled tests localized large errors to a
scale-sensitive eigensolver path, rather than QR. The explicitly labeled
stabilized variant forms the core and reconstructs factors in FP64,
uses CPU LAPACK FP64 EVD, and returns FP32 factors. It contains neither
FlashTSQR nor our custom Jacobi kernel. Native failing runs are archived
separately, not presented as evidence of an intrinsic FraQ quality loss.
All plotted methods pass relative action error on eight random probes below
$3\times10^{-5}$ and spectrum error below $10^{-4}$ over the full fleets;
stabilized FraQ's maximum action error is below $4.7\times10^{-7}$.

\paragraph{Recompression time and batching.}
FlashTSQR recompression takes 8.37\,s on Qwen50 and 8.08\,s on Llama35, versus
19.80\,s and 14.64\,s for stabilized FraQ: $2.37\times$ and $1.81\times$
speedups (Fig.~\ref{fig:conversion-latency}). SpecTraL takes 116.81/79.26\,s
and FLoRIST FP64 takes 281.28/191.06\,s on Qwen/Llama.
With identical FlashTSQR arithmetic but per-adapter batches capped at 24,
recompression instead takes 11.02/11.46\,s. Grouping across the fleet reduces
factor-construction calls from 596 to 26 for Qwen and from 385 to 17 for Llama,
yielding $1.32\times$/$1.42\times$ batching gains. Stabilized FraQ sees
no corresponding gain: its per-adapter times are 19.76/14.47\,s.
Thus, fewer dispatches alone do not guarantee faster recompression; the
operator implementation and feasible workspace sizes also matter.

\paragraph{Backend effects and interpretation.}
A diagnostic samples 24 real matrices from each of the 36 shape/rank groups.
Three-repeat stage medians, weighted by the fleets' group counts, attribute
about 74--76\% of SpecTraL's isolated stage time to GPU core SVD and
24--26\% to its two QRs. For rank-64 batches, GPU SVD takes roughly
193--200\,ms, versus about 17\,ms for CPU SVD including transfers; both
produce accurate cores. These sampled timings identify a bottleneck, not an
end-to-end CPU-SVD result. In particular, SpecTraL remains on GPU SVD while
stabilized FraQ uses CPU EVD. The measured gaps reflect algorithmic work,
precision, core backend, orientation, and scheduling together; they are not
isolated TSQR gains or platform-independent algorithmic ratios. A library
CPU/hybrid SVD control for SpecTraL remains unmeasured at full-fleet scale.

\subsection{Co-Resident Mixed-Fleet Serving on GH200}
\label{app:gh200-mixed-serving}
\begin{table}[!htbp]
\centering
\caption{\textbf{Mixed85 at 128 offered req/s.}
Five-seed means on one GH200; TTFT P50 is in seconds and TPOT P99 in milliseconds; Original and Spectral retain natural stopping.
Only Base is forced to the corresponding Spectral per-request output lengths.}
\label{tab:mixed85-gh200}
\footnotesize
\setlength{\tabcolsep}{3pt}
\begin{tabular}{lrrrr}
\toprule
Configuration & req/s & tok/s & TTFT & TPOT \\
\midrule
Original/native & 12.47 & 511 & 44.39 & 222.4 \\
Original/LoRAForge & 6.99 & 285 & 101.12 & 367.1 \\
Spectral/LoRAForge & \textbf{23.58} & \textbf{981} & \textbf{6.43} & \textbf{40.8} \\
Base (Spectral lengths) & 37.67 & 1,567 & 1.08 & 31.6 \\
\bottomrule
\end{tabular}
\end{table}

\paragraph{Protocol.}
One GH200 hosts two BF16 engines (three CPU threads and 16,384 KV tokens
each), with Qwen/Llama adapter budgets of 14/10\,GiB. Five seeds each
have 1,700 requests, 20 per adapter. All cases share prompts, routing and
Poisson arrivals. Original/Spectral retain natural EOS; only Base matches
Spectral's per-request output lengths. GH200 and LUMI use the same frozen
rank-floored $1/12$ spectral exports as quality evaluation. Comparisons remain
within platforms because execution backends and system behavior differ.
The custom arm uses the CUDA
implementation of RowExpand and our shrink kernel, together with effective-rank
metadata, compact shadow weights, and dispatch with CSGMV. Archived execution
records confirm entry into the custom prefill/decode expand and shrink paths;
the legacy counter names retain the \texttt{hip\_} prefix even on CUDA.

\paragraph{Observed gains.}
At 128 req/s, Spectral/LoRAForge reaches $1.89\times$/$1.92\times$
Original/native's request/output-token throughput; TTFT P50 falls from
44.39 to 6.43\,s and TPOT P99 from 222.4 to 40.8\,ms.
Against Original/LoRAForge, throughput ratios are $3.38\times$/$3.44\times$.
These system-level gains include compression, execution policy, residency
and queueing. Overload latency ratios are not isolated kernel speedups.

\paragraph{Residency evidence and attribution.}
Under identical byte budgets, native pool capacities are 28/15 slots for
Original/native, 16/9 for Original/LoRAForge, and 51/36 for Spectral/LoRAForge
(Qwen/Llama, including one no-adapter slot). Custom execution also stores
shadow weights. Only the compressed fleets fit entirely.
At 128 req/s, Original/native averages 848.4 adapter loads and 437.3\,GB
of H2D source bytes per run; Original/LoRAForge averages 1,005.2 loads and
493.0\,GB. Spectral/LoRAForge has zero timed adapter loads or H2D copy calls
at every main-sweep point. These software counters measure copy-source
bytes, not bus traffic or isolated transfer time. They demonstrate a
residency difference but do not quantify its contribution separately from
low-rank compute, custom kernels, output lengths or scheduling.
Evidence: \path{evidence/20261029/mixed85-gh200}.

\subsection{CTS Quality Baseline Protocol}
\label{app:cts-quality}

We use the unmodified \texttt{full\_lora\_pca} routine from the authors'
LoRALoLA revision \texttt{aab283c}, separately for every model, layer, and
projection. This is single-shared-basis JD-Full, without clustering.
For $n$ participating adapters and shared rank $r$, the parameter cost is
$r(d_{\mathrm{in}}+d_{\mathrm{out}})+nr^2$; missing target modules remain absent.
The largest common integer rank fitting each Uniform ceiling is
125/174/272/412 for Qwen and 113/165/273/432 for Llama, in budget order
$1/12,1/8,1/4,1/2$. Fits use original scaled updates, 100 iterations and seeds
101/202, selecting the lower raw squared Frobenius objective per module.
A four-module pilot at $1/12$ checked iteration counts through 300;
this does not establish convergence at every module or larger budget.

For the common native eager quality backend, we export equivalent independent
factors $A'_i=(A_iV)V^\top$ and $B'_i=U(U^\top B_i)$, with original LoRA
scaling absorbed into $B_i$ and unit export scaling. Their product is
$U\Sigma_iV^\top$ in exact arithmetic. Shared factors are retained in FP32;
evaluation exports use BF16, with algebraic and rounding checks recorded.
The expanded exports do not realize shared storage and are not a CTS serving
or throughput reproduction. All frozen token IDs, generation caps, natural
stopping and native scorers are reused. CTS initially uses four workers per
fleet; the remaining Qwen $1/2$ HypergraphFormer requests were redistributed
over eight workers, with 555 unique IDs verified after merging. Batch
composition can affect floating-point decoding.
Evidence: \path{evidence/20261030/cts-quality}.

\subsection{Per-Adapter Quality Regressions}
\label{app:quality-regressions}

Table~\ref{tab:quality-regressions} complements the adapter-macro scores in
Table~\ref{tab:quality-fleets} using the same full-evaluation protocol and
matched parameter budgets. The 21,592 frozen Qwen validation requests exclude
historical calibration IDs; the same request set is used across all quality arms.
For each compressed arm, $\Delta_{\min}$ is the
most negative adapter-score change relative to Original, and $N_{>5}$ counts
adapters losing more than five native-score points. These descriptive
statistics retain each task's native metric and do not pool examples across
tasks.

\begin{table}[ht]
\centering\fontsize{7.5}{9}\selectfont
\caption{\textbf{Per-adapter regressions across retained budgets.}
Each entry is $\Delta_{\min}/N_{>5}$ relative to Original; negative
$\Delta_{\min}$ indicates a score loss.}
\label{tab:quality-regressions}
\setlength{\tabcolsep}{0pt}
\renewcommand{\arraystretch}{1.35}
\begin{tabular}{@{}p{75pt}*{4}{>{\raggedleft\arraybackslash}p{\dimexpr(\linewidth-75pt)/4\relax}}@{}}
\toprule
Configuration & $1/2$ & $1/4$ & $1/8$ & $1/12$ \\
\midrule
\rowcolor{black!6}\multicolumn{5}{@{}l@{}}{\textbf{Qwen50}} \\
CTS & $-9.80/1$ & $-14.40/4$ & $-35.80/11$ & $-52.00/29$ \\
\textcolor{navy}{\lfuniform{}} & $-2.80/0$ & $-8.00/2$ & $-14.05/4$ & $-31.17/5$ \\
\textcolor{navy}{\lfspectral{}} & $-2.34/0$ & $-4.00/0$ & $-4.00/0$ & $-19.00/3$ \\
\lfspectral{} (w/o floor) & $-1.20/0$ & $-5.40/1$ & $-11.60/2$ & $-10.80/5$ \\
\midrule
\rowcolor{black!6}\multicolumn{5}{@{}l@{}}{\textbf{Llama35}} \\
CTS & $-8.62/3$ & $-9.09/6$ & $-23.80/9$ & $-24.10/9$ \\
\textcolor{navy}{\lfuniform{}} & $-1.11/0$ & $-2.28/0$ & $-4.91/0$ & $-8.60/6$ \\
\textcolor{navy}{\lfspectral{}} & $-1.25/0$ & $-1.41/0$ & $-3.47/0$ & $-7.51/1$ \\
\lfspectral{} (w/o floor) & $-1.40/0$ & $-7.60/2$ & $-7.60/3$ & $-7.60/4$ \\
\bottomrule
\end{tabular}

\end{table}

At $1/12$, the no-floor ablation's worst Qwen adapter loses 10.80 points, compared with 31.17
for \lfuniform{}; each has five adapters losing more than five points. Its worst Llama
loss is 7.60 points, with four adapters losing more than five, versus
\lfuniform{}'s 8.60 points and six adapters. \lfspectral{} with the fixed rank floor reduces these
Llama regressions to a 7.51-point worst loss and one adapter losing more than
five points. For Qwen at the same budget, however, the floor increases the
worst loss from 10.80 to 19.00 points even as the count falls from five to
three. Thus, improving one aggregate regression statistic need not improve
the others.

\subsection{Independent Single-Fleet Serving}
\label{app:single-fleet-serving}
\begin{figure*}[!t]
\centering
\begin{subfigure}{\textwidth}
\centering
\includegraphics[width=\linewidth,clip]{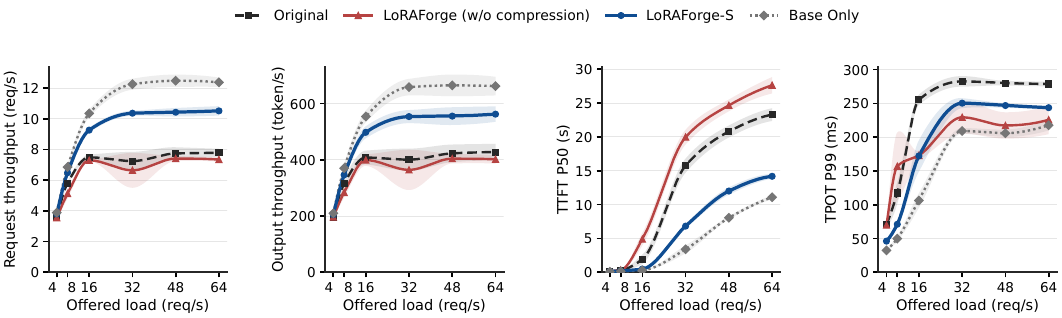}
\caption{Qwen3-4B-Instruct-2507: 50 adapters, 1,000 requests per seed and offered rate.}
\label{fig:serving-qwen}
\end{subfigure}
\par\smallskip
\begin{subfigure}{\textwidth}
\centering
\includegraphics[width=\linewidth,clip]{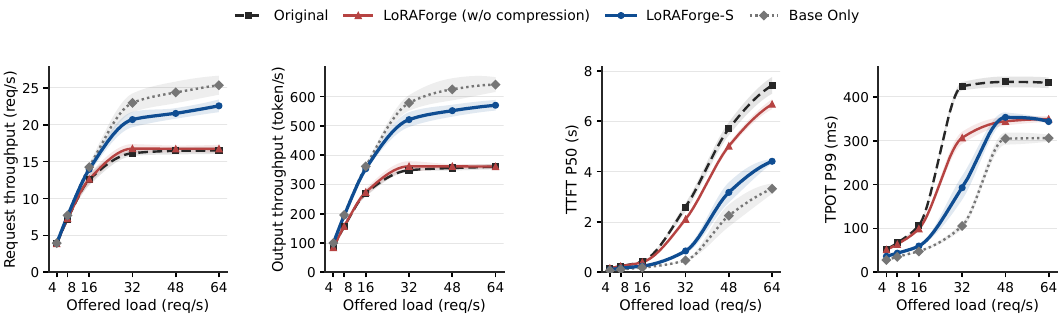}
\caption{Llama-3.1-8B-Instruct: 35 adapters, 700 requests per seed and offered rate.}
\label{fig:serving-llama}
\end{subfigure}
\caption{\textbf{Four-arm serving comparisons on two frozen adapter fleets.}
All six offered rates use five input seeds and the same prefill/decode interval of four.
Markers are measured means; bands are sample SD. Smooth connections are
shape-preserving guides, not additional measurements. LoRA outputs use natural
EOS or the task cap. Base replays frozen per-request Spectral lengths: the earlier
Qwen backend at 8/16/32, optimized Qwen at 4/48/64, and optimized Llama at all
rates. Natural-EOS throughput includes output-length and execution effects;
the two fleets also differ in model size and task mix.}
\label{fig:serving-current}
\end{figure*}

These measurements run Qwen50 and Llama35 separately on one MI250X GCD,
complementing the co-resident experiment in Figure~\ref{fig:mixed-platforms}.

\paragraph{Serving setup.}
Each run uses one MI250X GCD (64\,GB), BF16, tensor parallelism one,
and SGLang 0.5.22. We serve Qwen3-4B-Instruct-2507 with 50 pinned adapters
and Llama-3.1-8B-Instruct with 35 pinned adapters. The Llama study uses the same frozen rank-floored $1/12$ spectral exports
as Llama quality evaluation; Qwen uses its corresponding Qwen exports.
We disable radix/prefix reuse, use context length 16,384,
chunked prefill and maximum prefill tokens 8,192, maximum running requests 256,
a static memory fraction of 0.80, and the Triton attention backend.
Decode uses full graphs through batch 256. Qwen captures breakable prefill
graphs through 2,048 tokens, with stable normalization and replay eligibility
capped at 256 tokens. Llama retains native normalization and replays exact
prefill shapes of 4/8/16/32/64/128/256 tokens; other shapes run eagerly.
The rank-aware dispatch policy is transferred unchanged from Qwen to Llama.
The prefill/decode interval is fixed at four for all systems, so this scheduler
setting is not a claimed optimization.

For each offered rate in $\{4,8,16,32,48,64\}$ requests/s, we use five frozen input
samples (seeds 101--105), each containing 20 requests per adapter:
1,000 requests for Qwen and 700 for Llama.
Input identities/order, task caps, and seeded Poisson arrival schedules are
matched across systems. Runs include correctness checks, warmup, and an untimed
conditioning pass. LoRA requests stop naturally at EOS or their task cap.
Qwen's 4/48/64-rate comparisons rerun both uncompressed backends;
8/16/32 use their earlier completed five-seed references. Llama's 48-rate
comparison is a separate supplement to its completed 4/8/16/32/64 sweep.
All plotted points have five completed seeds; failed timing attempts are excluded.
These are samples
from the frozen evaluation pools, not newly collected held-out quality sets.

\begin{table}[t]
\centering\footnotesize
\caption{\textbf{At 64 offered requests/s: five-seed means.}
All 1,000-request runs complete without failed requests. Base exactly matches
the optimized Spectral's per-request output lengths at this offered rate.}
\label{tab:serving64}
\setlength{\tabcolsep}{3.5pt}
\renewcommand{\arraystretch}{1.18}
\begin{tabular}{@{}lrrrr@{}}
\toprule
\rowcolor{resulthead} & Output & Request & TTFT & TPOT\\
\rowcolor{resulthead}System & tok/s & req/s & P50 (s) & P99 (ms)\\
\midrule
{\color{black!60}Base, length replay} & {\color{black!60}661.8} & {\color{black!60}12.37} & {\color{black!60}11.10} & {\color{black!60}217.6}\\
Original / native & 427.3 & 7.78 & 23.30 & 278.6\\
Original / optimized & 401.5 & 7.34 & 27.63 & 225.8\\
\rowcolor{resultours}\textbf{Spectral / LoRAForge} & 562.2 & 10.51 & 14.18 & 243.6\\
\bottomrule
\end{tabular}
\end{table}

\paragraph{End-to-end gains and trade-offs.}
At 64 offered requests/s, Spectral with the optimized path reaches 562.2 tok/s
versus 427.3 for uncompressed native CSGMV: $1.32\times$ token throughput and
$1.35\times$ request throughput (Table~\ref{tab:serving64}). It reaches 84.9\%
of the matched-output Base reference. At 8/16/32, the corresponding token
throughput gains over the archived uncompressed-native references are
$1.10\times$, $1.22\times$, and $1.38\times$.
The systems are overloaded relative to their achieved request rates at the
higher offered loads; these measurements do not establish that they sustain
64 requests/s or satisfy a production latency SLO.

At 48 offered requests/s, Qwen Spectral reaches 555.2 tok/s versus native's
422.9, a $1.31\times$ token-throughput ratio and $1.35\times$ request-throughput
ratio. The Qwen 32-rate dip is retained: in seed 102, one HypergraphFormer
request repeats tree identifiers until its 3,328-token cap, compared with about
1,270 tokens at 48/64. Another optimized seed also executes more slowly
without a longer output. All requests complete with valid stream termination;
these observations do not establish that performance must decrease at that load.

\paragraph{Transfer to the Llama fleet.}
Figure~\ref{fig:serving-llama} applies the frozen dispatch policy to a second
model and adapter cohort. At 64 offered requests/s, Spectral reaches 570.4 tok/s
and 22.56 completed requests/s, versus native's 360.6 tok/s and 16.51 requests/s:
$1.58\times$ token throughput and $1.37\times$ request throughput.
Mean output length also increases from 21.84 to 25.29 tokens, so the token ratio
is not a fixed-work hardware speedup. At 48, the corresponding token throughputs
are 551.9 and 356.2 tok/s. Table~\ref{tab:quality-fleets} reports full task
quality for the same adapter cohorts and the same rank-floored spectral
exports at $1/12$.
Table~\ref{tab:fixedwork64} additionally reports a provisional fixed-work
Llama control with two complete paired seeds and common KV capacity.

\subsection{Evidence Boundaries and Remaining Measurements}
\label{app:evidence-boundaries}
Spectral energy does not bound task accuracy or enforce tenant fairness.
The full quality sweeps cover both policies on the two served cohorts, using
rank-floored spectral exports shared with deployment at $1/12$. They use exposed
public evaluation pools and do not establish statistical equivalence, unseen-task
generalization, or universal tenant fairness. Budgets model parameters,
not latency, popularity, or output length. Our co-design couples representation
and execution; it does not jointly search ranks, layouts, and kernels under a
quality--latency objective. The controls establish the benefit and scope of the
tested execution policy, not optimality of the combined design. The crossed
control covers one Qwen budget and two offered loads; three input seeds do not
establish broad workload generality. Both policies benefit from the shared
runtime, while Uniform is faster and has a lower allocated-memory peak in this
control. These are measured tradeoffs, not a quality--throughput frontier.
The main dispatch policy is specialized to the tested MI250X/software
configuration. The CUDA implementation is also evaluated on GH200, but these
two platforms do not establish portability of performance across arbitrary
GPUs. The native pool remains beside the compact $B$ shadow on both platforms;
the shadow adds resident storage rather than replacing the pool. Compact
parameter counts therefore do not describe total resident LoRA memory.
Natural-EOS runs combine execution and behavioral effects; fixed-output controls
isolate backend throughput when both compared paths share the same stopping
contract, not task equivalence. Output-length matching alone does not control
admission or batching, as the co-resident Base diagnostic illustrates.
Recompression timings measure full-factor operator passes. The main GH200
control isolates QR replacement within a shared pipeline, but does not isolate
every FlashTSQR kernel contribution. The supplementary MI250X comparison
uses different core backends and batching schedules; its ratios combine
these effects. Full-fleet
CPU/hybrid SVD controls, end-to-end loading/export cost, and amortization over
serving lifetime remain unmeasured.

\end{document}